%% file: PTQTP.tex
\documentclass[11pt]{article}

\usepackage[preprint]{acl}

\usepackage{times}
\usepackage{latexsym}

\usepackage[T1]{fontenc}
\usepackage[utf8]{inputenc}

\usepackage{microtype}

\usepackage{inconsolata}

\usepackage{hyperref}       % hyperlinks
\usepackage{url}            % simple URL typesetting
\usepackage{booktabs, siunitx}       % professional-quality tables
\usepackage{multirow} % for multirow cells in tables
\usepackage{amsfonts}       % blackboard math symbols
\usepackage{algorithm}      % for algorithm environment
\usepackage{nicefrac}       % compact symbols for 1/2, etc.

\usepackage{graphicx}

\usepackage{algpseudocode}
\usepackage{array, multirow, bold-extra, caption}
\usepackage{arydshln}
\usepackage{xcolor}      % 颜色支持
\usepackage{algorithm}
\definecolor{checkgreen}{RGB}{0,128,0}  % 深绿色
\definecolor{crossred}{RGB}{200,0,0}    % 暗红色

\usepackage{booktabs}
\usepackage{enumitem}
\usepackage{subfigure}  %插入多图时用子图显示的宏包

\usepackage{pgf}           % For \pgfmathsetmacro
\usepackage{booktabs}      % For table rules
\usepackage{amsmath}       % For \textsubscript
\usepackage{graphicx}      % For \resizebox
\usepackage{tablecolor}
\usepackage{amsthm}     % 定理环境核心支持
\usepackage{amssymb}    % 数学符号扩展(包含\mathbb)
\usepackage{mathtools}  % 符号对齐与高级数学工具
\usepackage{bm}         % 粗体数学符号支持
\theoremstyle{definition}

\usepackage{mdframed}

\title{PTQTP: Post-Training Quantization to Trit-Planes for Large Language Models}

\author{
 \textbf{He Xiao\textsuperscript{1}\footnote[1]{}},
 \textbf{Runming Yang\textsuperscript{1}\footnote[1]{}},
 \textbf{Qingyao Yang\textsuperscript{1}\footnote[1]{}},
 \textbf{Wendong Xu\textsuperscript{1}},
\\
 \textbf{Zhen Li\textsuperscript{2}},
 \textbf{Yupeng Su\textsuperscript{3}},
 \textbf{Zhengwu Liu\textsuperscript{1}},
 \textbf{Hongxia Yang\textsuperscript{2}},
 \textbf{Ngai Wong\textsuperscript{1}\footnote[2]{}}
\\
 \textsuperscript{1} The University of Hong Kong
 \textsuperscript{2} The Hong Kong Polytechnic University\\
 \textsuperscript{3} University of California, Santa Barbara
\\
 \small{
   \footnote[1]{} Equal Contribution
   \footnote[2]{} Corresponding author: \href{mailto:email@domain}{nwong@eee.hku.hk}
 }
}

\begin{document}
\maketitle
\begin{abstract}

Post-training quantization (PTQ) of large language models (LLMs) to extremely low bit-widths remains challenging due to the fundamental trade-off between computational efficiency and representational capacity. While existing ultra-low-bit methods rely on binary approximations or quantization-aware training(QAT), they often suffer from either limited representational capacity or huge training resource overhead. We introduce \textbf{PTQ} to \textbf{T}rit-\textbf{P}lanes (PTQTP), a structured PTQ framework that decomposes weight matrices into dual ternary \(\{-1, 0, 1\}\) trit-planes. This approach achieves multiplication-free additive inference by decoupling weights into discrete topology (trit-planes) and continuous magnitude (scales), effectively enabling high-fidelity sparse approximation. PTQTP provides: (1) a theoretically grounded progressive approximation algorithm ensuring global weight consistency; (2) model-agnostic deployment without architectural modifications; and (3) uniform ternary operations that eliminate mixed-precision overhead. Comprehensive experiments on LLaMA3.x and Qwen3 (0.6B-70B) demonstrate that PTQTP significantly outperforms sub-4bit PTQ methods on both language reasoning tasks and mathematical reasoning as well as coding. PTQTP rivals the 1.58-bit QAT performance while requiring only single-hour quantization compared to 10-14 GPU days for training-based methods, and the end-to-end inference speed achieves 4.63$\times$ faster than the FP16 baseline model, establishing a new and practical solution for efficient LLM deployment in resource-constrained environments.
Code will available at https://github.com/HeXiao-55/PTQTP.

\end{abstract}

\section{Introduction}
\begin{figure}[t]
  \centering 
  \includegraphics[width=0.9\linewidth]{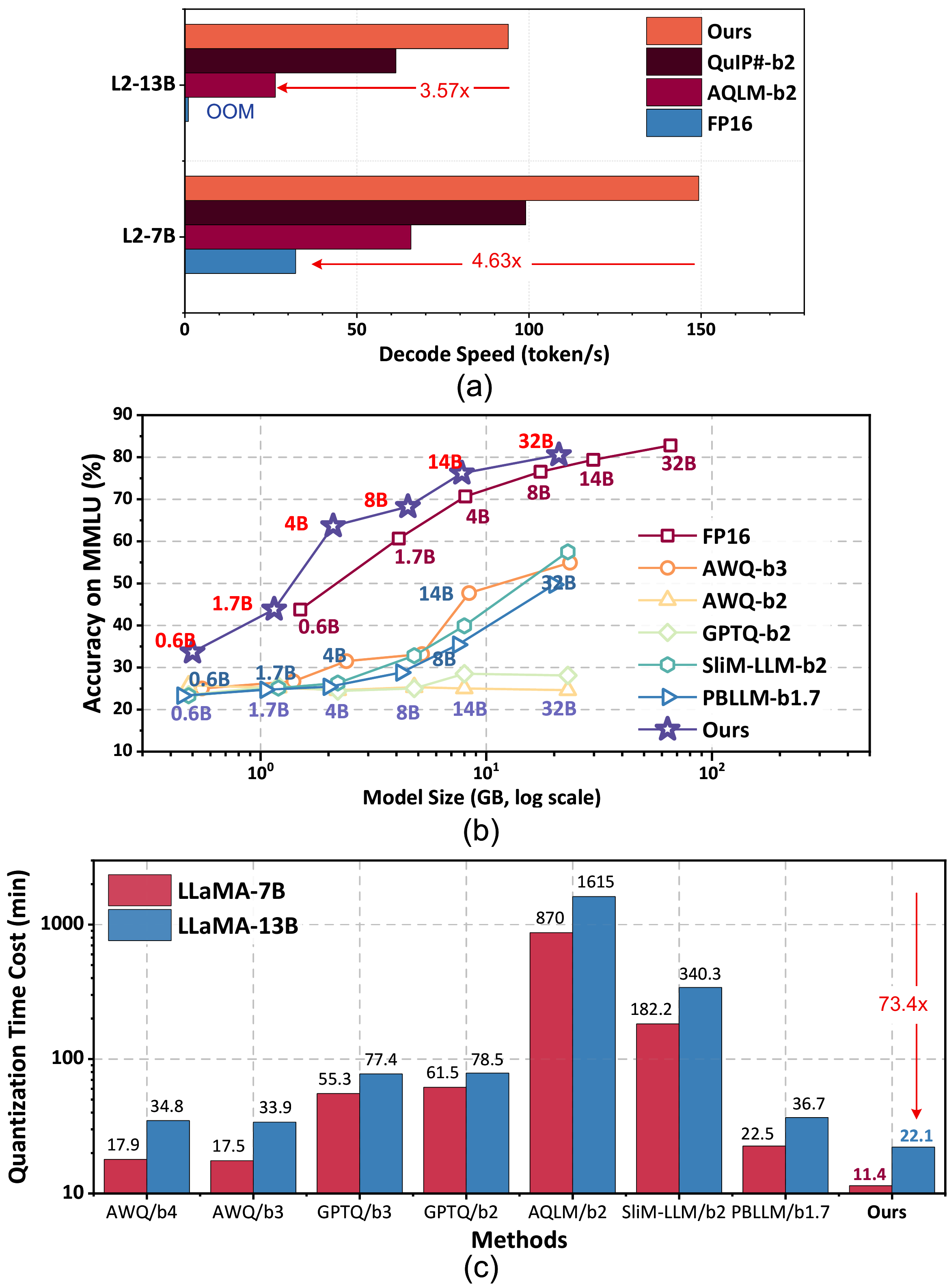} 
  \vspace{-1em}
  \caption{PTQTP minimizes compression costs while maintaining excellent performance. (a) PTQTP can deliver 4.63$\times$ end-to-end decode speed on NVIDIA RTX 3090 GPU. (b) The language reasoning evaluation on Qwen3 demonstrates that PTQTP outperforms existing low-bit quantization methods. (c) Quantization runtime comparison shows PTQTP achieves 73.4$\times$ speedup over AQLM and 1.57$\times$ over AWQ on LLaMA-7B.}
  \label{fig0} 
  % \vspace{-0.5em}
\end{figure}
\label{sec:intro}
The unprecedented success of large language models (LLMs) has revolutionized natural language processing, but their deployment is hindered by exorbitant computational and memory demands~\citep{brown2020language}. Models such as LLaMA3~\citep{llama3}, Qwen3~\citep{qwen3}, DeepSeek-R1-671B~\citep{guo2025deepseekr1}, with hundreds of billions of parameters, require specialized hardware and massive energy consumption, limiting accessibility on edge devices and raising environmental concerns~\citep{patterson2021carbon}. To address this, sub-4 bit low-bit quantization including binary (1-bit) and ternary (1.58-bit), has emerged as a viable approach, facilitating affordable memory consumption aligned with efficient and edge implementation demands.

Post-training quantization (PTQ) offers a practical pathway for compressing pretrained LLMs without retraining, with recent advancements like GPTQ~\citep{frantar-gptq}, AWQ~\citep{jlin2024AWQ}, and many exquisite PTQ methods ~\citep{shao2023omniquant,ashkboos2024quarot,xiao2023smoothquant,tseng2024qtip} achieving effective 4-bit quantization with near full-precision accuracy; Near-2bit PTQ methods~\citep{aqlm,huang2024slim,lieq,quip1,chee2023quip,zhao2025ptq1.61} concentrated on solving the outlier challenge by using pre-processing or high-precision protection methods before PTQ to smooth outliers and extra fine-tuning strategies to help refine the performance of the quantized model.

However, 2-4 bit operations still rely on costly multiply-accumulate (MAC) operations on existing hardware. Binary (1-bit) and ternary (1.58-bit) PTQ can effectively reduce algebraic multiplication to addition to save the inference consumption, but push PTQ to extreme bit-widths coexist with challenges and opportunities. \textit{\textbf{Two main Challenges}}: Binary PTQ that achieves 1-bit quantization through unstructured weight categorization, but sacrifices representational capacity~\citep{huang2024billm,li2025arbllm,shang2023pbllm,zhao2025ptq1.61}; Quantization-aware training (QAT) with extremely low-bit widths, which requires costly retraining, e.g., BitNet~\citep{BitNet,ma2024bitnet1.58} for pretraining binary/ternary models.\textit{ \textbf{Opportunities for Ternary}}: Ternary operations further enhance the logic selection capabilities compared to binary operations. Moreover, from the hardware perspective, ternary PTQ is not merely quantization but a step towards transforming LLM from an \textit{arithmetic-bound} to a \textit{logic-bound} operation. This represents a qualitative leap for future neuromorphic computing or edge AI. Structured ternary PTQ, offering higher expressivity than binary while avoiding QAT's pretraining overhead and extra fine-tuning, remains underexplored (if not unexplored). 

We introduce PTQTP (\textbf{P}ost-\textbf{T}raining \textbf{Q}uantization to \textbf{T}rit-\textbf{P}lanes), a novel structured PTQ method that bridges the gap between binary efficiency and ternary expressiveness while avoiding costly training overhead and time-consumed fine-tuning step. Unlike binary quantization, which \textit{forces weights to $\pm 1$, introducing significant quantization noise that disrupts precise logic flows}, causing sophisticated reasoning (such as mathematics) to collapse. PTQTP leverages a magnitude-topology decoupling strategy, it decomposes full-precision weights into a linear superposition of dual ternary trit-planes $\{-1, 0, 1\}$ modulated by a column of row-wise scalars. This essentially performs a constructive interference: the ``0'' state allows the model to selectively silence noise features. Furthermore, by converting heavy multiplications into lightweight ternary additions, PTQTP offers a practical, uniform, and mathematically robust solution for the deployment of powerful LLMs, opening new frontiers for efficient inference in real-world applications. 

1. PTQTP captures both the principal skeleton and the residual details of the weight distribution using a collaborative dual trit-planes structure, producing an expressive space that is more flexible and contains richer sparsity than current 1-3 bit methods. Our work demonstrates that structured ternary quantization strikes a sweet spot between computational simplicity and representational power, as shown in Fig.\ref{fig0}. 

2. PTQTP preserves complex inference topology at extremely low-bit precision, addressing a critical reasoning performance degradation in the literature. PTQTP offers a robust, model-agnostic solution for extremely low-bit PTQ. By eliminating the need for architecture-specific adjustments, it supports seamless deployment on quantization-sensitive models (e.g., LLaMA3.x, Qwen3), while consistently preserving performance and surpassing state-of-the-art architecture-dependent approaches in scalability. 

3. Extensive experiments demonstrate that PTQTP consistently outperforms state-of-the-art low-bit methods, surpassing most 1–3 bit PTQ approaches in language benchmarks while enabling faster quantization and hardware-efficient multiplication-free operations. Remarkably, it rivals or even exceeds 1.58-bit QAT, despite requiring over $10^4\times$ fewer GPU hours, without any retraining or post-PTQ fine-tuning, underscoring its efficiency and generalizability across advanced models.

\begin{figure*}
  \centering 
  \includegraphics[width=0.7\textwidth]{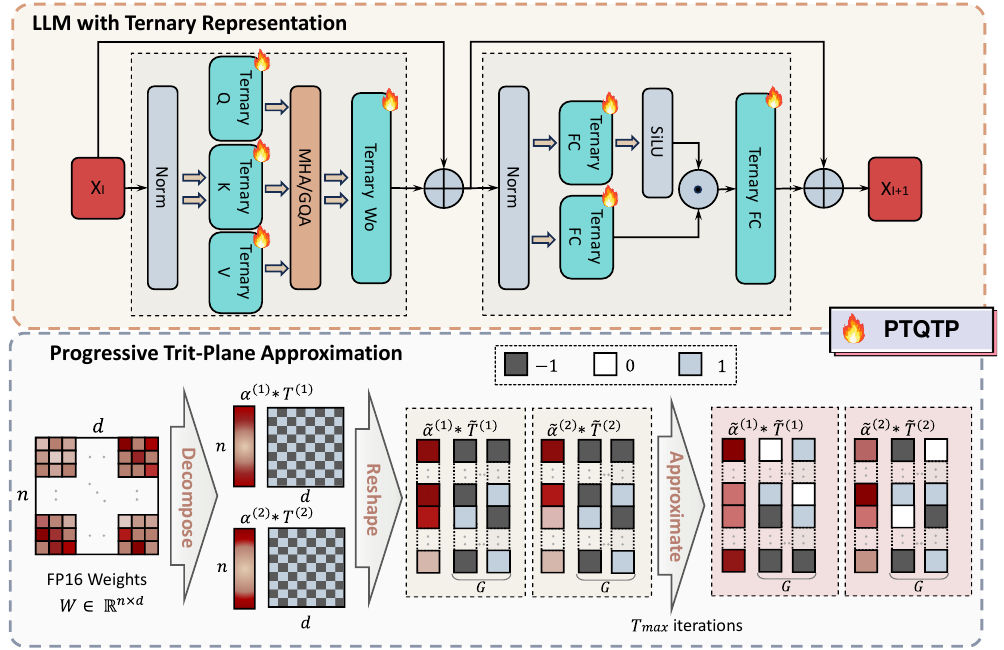} 
\caption{PTQTP workflow overview: (top) Linear layer transformation pathway for ternary quantization in LLaMA architecture; (bottom) Group-wise progressive trit-plane approximation process, where $G$ represents group size and $T_{max}$ indicates maximum iteration count.}
  \label{fig1} 
  % \vspace{-1em}
\end{figure*}

\section{Methodology}
\label{sec:methods}
% \subsection{Background}

\subsection{Magnitude-Topology Decoupling}
Unlike conventional scalar quantization which maps weights directly to a discrete grid, we conceptualize the quantization of LLM weights as a magnitude-topology decoupling problem.
Let $W \in \mathbb{R}^{n \times d}$ be the weight matrix. We assume that $W$ carries two distinct types of information: (1)~\textit{Magnitude:} The amplitude or energy of these connections, which varies across channels. (2)~\textit{Topology:} The structural connectivity indicating which input features positively or negatively contribute to the output. 

Existing binary quantization $W \in \{-1, +1\}$ suffers from the \textit{forced activation} problem, where near-zero weights, \textit{i.e., noise}, are amplified to $\pm 1$, destroying the delicate logical inference paths essential for \textbf{complex reasoning tasks like \textit{MATH} and \textit{Coding}}. PTQTP addresses this by decomposing $W$ into a linear superposition of dual sparse trit-planes:
\begin{align}
    W &\approx \hat{W} = \sum_{k=1}^{K=2} \underbrace{\text{diag}(\alpha^{(k)})}_{\textit{Magnitude}} \cdot \underbrace{T^{(k)}}_{\textit{Topology}} \\
     W &\approx \underbrace{\alpha^{(1)} T^{(1)}}_{\textit{Coarse Structure}} + \underbrace{\alpha^{(2)} T^{(2)}}_{\textit{Fine-grained Correction}} 
\end{align}
Where $T^{(k)} \in \{-1, 0, 1\}^{n \times d}$ represents the discrete routing topology, and $\alpha^{(k)} \in \mathbb{R}^n$ represents the continuous channel gain.
Crucially, the inclusion of the ``0'' state allows PTQTP to perform implicit \textit{denoising}, filtering out irrelevant features rather than forcing them into binary states. 
The superposition of dual trit-planes enables \textit{constructive interference}, where the second plane $T^{(2)}$ acts as a spectral residual compensator, capturing the high-frequency details missed by the principal structural plane $T^{(1)}$ while ignore the parts is reconstructed.

\subsection{Dual Trit-planes Approximation}
The general process of PTQTP is illustrated in Fig.\ref{fig1}. We use \( T_i^{(k)} \) to denote the $i$th row of the \( k \)th trit-plane. First, we fix $T_i^{(k)}$ and optimize the scaling coefficients $\alpha_i^{(k)}$ without any bias terms. To solve for optimal scaling coefficients $\alpha_i^{(k)}$, we use the adaptive ridge regression, i.e., linear regression with a regularization term to the least-squares function.  We first define the local basis matrix $S_i$:
\begin{align}
    S_i &= \left[(T^{(1)}_i)^T~(T^{(2)}_i)^T\right] \in \{-1,0,1\}^{d \times 2} \\ 
    A_{i} &= (S_{i})^T S_{i} + \lambda_{i} I_2 \in\mathbb{R}^{2\times 2} \\
    b_{i} &= (S_{i})^T W_{i}^T\in\mathbb{R}^{2}
    \label{eqn:ridge}   
\end{align}

\begin{figure}[tbp]
	\centering
	\includegraphics[width=\linewidth]{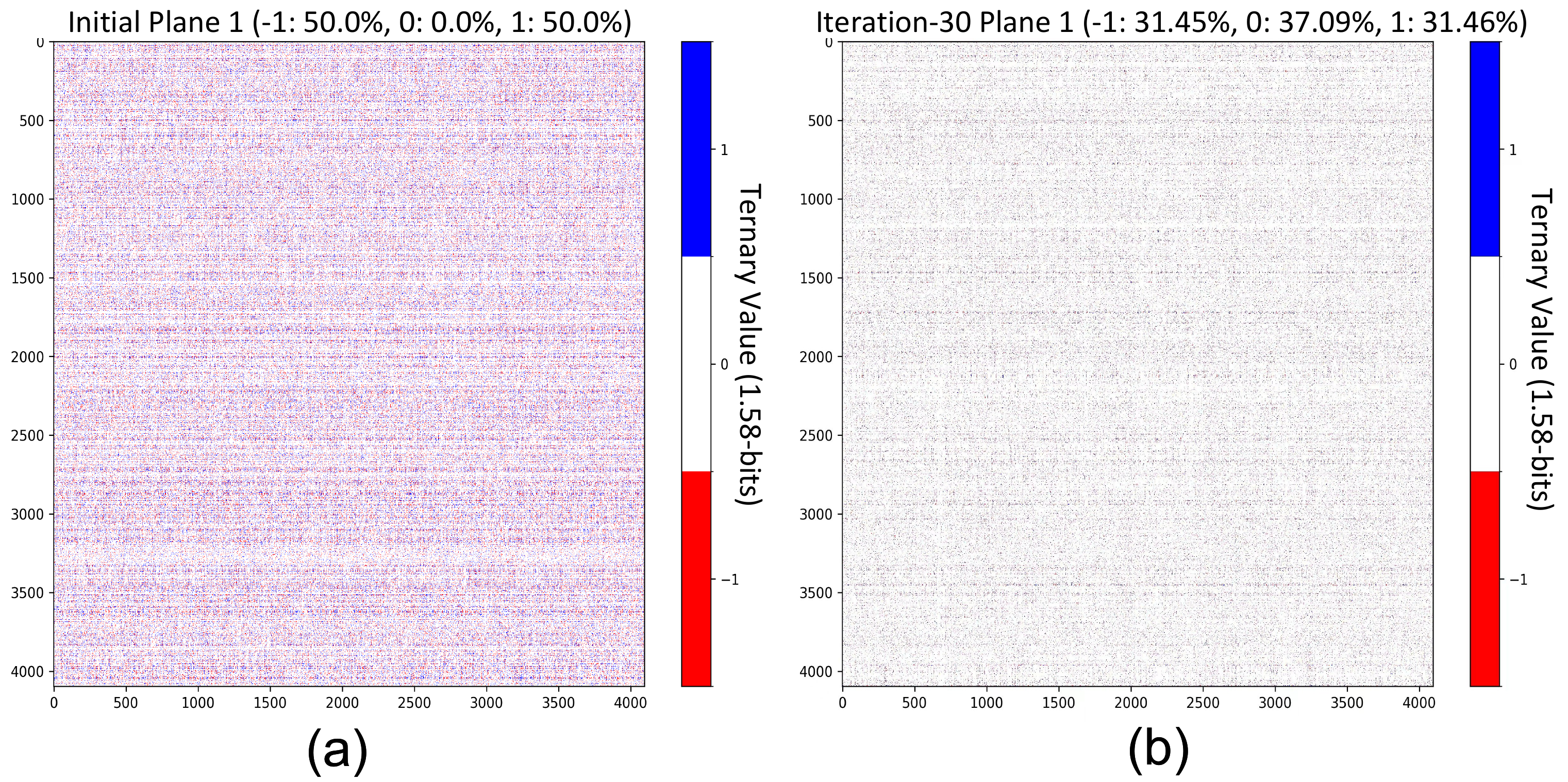}
	\caption{Iterations on LLaMA3.1-8B. (a) Initialized single trit-plane. (b)After 30 iterations.}
	\label{fig:ternary}
    % \vspace{-0.5em}
\end{figure}

Here $\lambda_i \in \mathbb{R}$ is a regularization parameter to improve numerical stability, \(I_2\) is the \(2\times2\) identity matrix. By constructing the local basis matrix row by row, we can find a closed-form solution $\alpha_i (\in\mathbb{R}^2):=[\alpha^{(1)}_i \alpha^{(2)}_i]^T=A_{i}^{-1}b_{i}$ for each $i$ independently. However, due to the discrete nature of trit-planes, the method may converge to different minima depending on the regularization parameter \(\lambda_{i}\), which is crucial for the regression performance: if it is too small, the regularization effect may be negligible and the solution may be unstable. If it is too large, the coefficients \(\alpha_i\) can be overly diminished, leading to a poor approximation of the original weight matrix.

\subsection{Progressive Optimization}

\paragraph{Adaptive Regularization.}
To achieve a robust decomposes process, we adaptively regularize the approximation and further optimize it through progressive optimization. First, we estimate the condition number of the $2\times 2$ system:
\begin{equation}
\kappa_{i,approx} = \|A_{i}\|_F \cdot \|A_{i}^{-1}\|_F
\end{equation}
This measure guides dynamic updates to \(\lambda_{i}\) with the constraint \(\lambda_{i,new} = \lambda_{i} \leq \lambda_{\text{max}} (:= 1.0)\) when $\kappa_{i,approx} < 10^{6}$.
% This measure guides dynamic updates to \(\lambda_{i}\) with the constraint \(\lambda_{i,new} \leq \lambda_{\text{max}} (:= 1.0)\) :
% \begin{align}
% \lambda_{i,new} = 
% \begin{cases}
% \lambda_{i} ,  \quad\kappa_{i,approx} < 10^{6} \\
% \lambda_{i}  \sqrt{{\kappa_{i,approx}/{10^{6}}}}, \quad \text{otherwise}
% \end{cases}\label{eqn:lamda_i}
% \end{align}
This adaptation mitigates under-regularization (leading to singularity and unstable solutions) and over-regularization (causing excessive coefficient shrinkage), ensuring robustness across weight blocks with varying numerical conditions. Furthermore, the optimal scaling coefficients $\alpha_i$ are then found by solving the following optimization problem:
\begin{equation}
    \alpha_i = \mathop{\arg\min}_{\theta_i \in \mathbb{R}^2} \left\|W_{i}-S_{i}\theta_i \right\|_F^2 + \lambda_{i}\|\theta_i\|_F^2 \label{eq:theta}
\end{equation}%

Where $\theta_{i}$ is a crucial variable that represents the intermediate solution to the scaling coefficients in the process of updating $\alpha_i$. In addition, \(\lambda_{i}\|\theta_i\|_F^2\) is the regularization term that penalizes large values of coefficients \(\theta_i\). The Frobenius norm term represents the squared error between the linear combination of the \(W_i\) and $\hat{W}_i$, where \(S_{i}\) incorporates the current iteration's trit-plane values. 
% We adapt \(\lambda_i\) using Eq. \eqref{eqn:lamda_i}, aligning regularization strength with local weight properties. 
Specifically, we perform the optimization steps $\leq T_{max}$. Each update step is designed to not increase the Frobenius norm \( \| W - \hat{W} \|_F^2 \). By ensuring that this norm monotonically decreases, we can guarantee that the algorithm will converge to a local minimum. Therefore, we refine trit-plane elements through local exhaustive search, updating \(T_{i,j}^{(k)}\) to the value in \(c_m \in \{-1, 0, 1\}\) that minimizes squared error:  
\begin{equation}
T_{i,j}^{(k)} = \mathop{\arg\min}_{c_m^{(k)} \in \{-1,0,1\}} \left(W_{ij} - \sum_{k=1}^2 \alpha_i^{(k)} c_m^{(k)} \right)^2
\end{equation}
% Unlike the unstructured mask used for salient weights search in~\citep{huang2024billm,li2025arbllm}, 
Therefore, PTQTP achieves \(\mathbb{E}[W_{ij}] \approx \alpha_i^T \cdot \mathbb{E}\left[[T^{(1)}_{ij}~T^{(2)}_{ij}]^T\right]\) and is \textit{bias-free} and \textit{mask-free}, its uniform model architecture preserves hardware-friendly design and operations. Fig. \ref{fig:ternary} illustrates the example process of a single trit-plane update.

\input{algs/app_group_algro}

\input{tables/main_results}

\paragraph{Batch Processing.}
We further introduce group-wise (aka block-wise) processing~\citep{frantar-gptq,jlin2024AWQ} in PTQTP, namely, by reshaping $W$ from $n\times d$ to $\frac{nd}{G}\times G$, or equivalently grouping into $G$ columns. Specifically, we set $G=128$ similar to other related works, and we denote such group-wise operation with the tilde $\tilde{(\circ)}$ notation, then we have
\begin{align}
    \tilde{A}_i = \tilde{S}_i^T &\tilde{S}_i + \lambda_i I_2, \\ 
    \tilde{\alpha}_i = \tilde{A}_i^{-1}\tilde{b}_i, &\quad
    \tilde{b}_i = \tilde{S}_i^T \tilde{W}_i^T 
\end{align}
Such grouping of columns generally reshapes $W$ into a taller $\tilde{W}$, alongside with lengthened $\tilde{\alpha}^{(k)}$'s. Nonetheless, the increase in the latter incurs negligible parameter overhead compared to the model size, which is far outweighed by an improved approximation and performance. 

As shown in \textbf{Algorithm \ref{alg_group}}, we first decompose all the FP16 linear projection weights into trit-planes for magnitude-topology decoupling and approximation. Then we apply progressive optimization and adaptive regulation to automatically save optimal parameters. Furthermore, the unimportant information is filtered out, while the salient feature information is retained. \textit{Experiment results demonstrate that PTQTP always converges within 50 iterations}, enabling a stable and efficient compression method.

\section{Experiments}
% \subsection{Settings}

\paragraph{Quantization Configuration.}
We implemented PTQTP on PyTorch platform using models from Huggingface~\citep{hf}. All the weights in linear projection were quantized with tolerance $\epsilon = 10^{-4}$ and maximum iterations $T_{max}=50$. For numerical stability, we employed dynamic regularization with $\lambda \in [10^{-8}, 1]$. No task-specific calibration, tuning, or fine-tuning was applied in any experiment. All evaluations were conducted on a single NVIDIA A100 80GB GPU.

\paragraph{Model Architectures.}
We evaluated PTQTP across multiple mainstream LLM families including Qwen3~\citep{qwen3}, LLaMA3.x~\citep{llama3} and LLaMA series~\citep{touvron2023llama, touvron2023llama2openfoundation}. To assess cross-domain generalization capabilities, we also tested instruction-tuned variants of these models.

\paragraph{Baseline Methods.} 
We compared PTQTP with three categories of methods: (1) Extremely low-bit PTQ methods including: PBLLM~\citep{shang2023pbllm}, SliM-LLM~\citep{huang2024billm}, QUIP\#~\citep{quip1}, QUIP~\citep{chee2023quip}, AQLM~\citep{aqlm}; (2) Popular PTQ Methods including: GPTQ~\citep{frantar-gptq}, AWQ~\citep{jlin2024AWQ}, QTIP~\citep{tseng2024qtip}, OmniQuant~\citep{shao2023omniquant}; and (3) 1.58-bit QAT approaches~\citep{ma2024bitnet1.58, ma2025bitnetb1582b4t} to demonstrate PTQTP's effectiveness even against quantization-aware training-based methods. For fair comparison with smaller models, we included common 1B-3B LLMs such as SmolLM2~\citep{allal2025smollm2} and MiniCPM~\citep{hu2024minicpm}.

\input{tables/New_table2_L2Q3}

\paragraph{Evaluation Protocol.}
We assessed language modeling capability through perplexity measurements on WikiText-2~\citep{wiki} and C4~\citep{c4}. To evaluate reasoning abilities, we utilized ARC-Challenge, ARC-Easy~\citep{arc-c}, BoolQ~\citep{boolq}, HellaSwag~\citep{hellaswag}, PIQA~\citep{piqa}, Winogrande~\citep{winog}, and MMLU~\citep{mmlu}. For comprehensive comparison with SOTA methods, we further evaluated coding and mathematical reasoning performance on LLaMA3.x and Qwen3 models. All evaluations were conducted using the standard lm-eval benchmarking tool~\citep{eval-harness}.

\subsection{Main Results}

\paragraph{Perplexity on Mainstream Model Backbones.}
Table \ref{tab_qwen} compares the perplexity of PTQTP on WikiText2 and C4 datasets with baselines across LLaMA2 , LLaMA3.x and Qwen 3 variants (0.6B to 70B). The results demonstrate that PTQTP consistently outperforms existing extremely low-bit (1-3 bit) quantization schemes and approaches or exceeds 4-bit methods across diverse architectures. \textbf{\textit{This robustness is particularly pronounced in SOTA small LLMs (0.6B-3B), which are typically more vulnerable to quantization due to their higher information density from advanced pretraining recipes.}} The exceptional performance retention of PTQTP establishes it as a robust, generalizable, and efficient quantization solution for both current and future models.

\input{tables/newtable3_benchmarkQAT}
\paragraph{Language Reasoning Tasks.} 
Models with larger parameters generally exhibit greater resistance to quantization-induced performance loss, making them ideal for evaluating extreme quantization effectiveness. Results in Table~\ref{tab_zero_shot_results} reveal dramatic disparities in capability retention across PTQ methods. Existing solutions exhibit significant performance degradation when performing extremely low bit quantization (less than 2 bits), with a performance gap of almost half. In contrast, PTQTP demonstrates its consistent performance across diverse benchmarks (\textbf{\textit{average performance accuracy retention near to 95\%, and the language reasoning ability is very close to 4-bit methods though PTQTP only use ternary representation precision}}) without requiring dynamic bit allocation, salient weight protection, or sensitivity-aware quantization fundamentally challenges the trade-off between quantization and reasoning ability.

\paragraph{Complex Reasoning \textit{vs} FP16/1.58-bit QAT Language Models.}
PTQTP's versatility enables extremely low-bit quantization of advanced models while maintaining near-baseline performance. Unlike QAT methods that require extensive pretraining and fine-tuning, PTQTP applies uniform treatment across model layers without post-quantization adjustments. Table~\ref{tab_compare_qat} demonstrates PTQTP's minimal performance degradation compared to baselines, and its ability to match or exceed 1.58-bit QAT BitNet schemes~\citep{ma2024bitnet1.58,ma2025bitnetb1582b4t} at similar model sizes. \textbf{\textit{These results confirm PTQTP's exceptional stability and establish it as a true plug-and-play solution for model-agnostic 1.58-bit quantization.}}
Furthermore, compared to other existing PTQ schemes in extreme cases (2 bits), we found that they also perform poorly in terms of mathematical capabilities, especially without using an additional codebook, such as AWQ, which almost completely loses its mathematical and coding capabilities. On the other hand, PTQTP can better protects overall performance including complex reasoning, even compared to the AQLM (with additional codebook).

\subsection{Ablation Studies} 
\begin{figure}[t]
	\centering
	% \subfigbottomskip=2pt
	\subfigcapskip=2pt
	\begin{tabular}{cccccc}
		\subfigure[]{
			\includegraphics[width=0.45\linewidth]{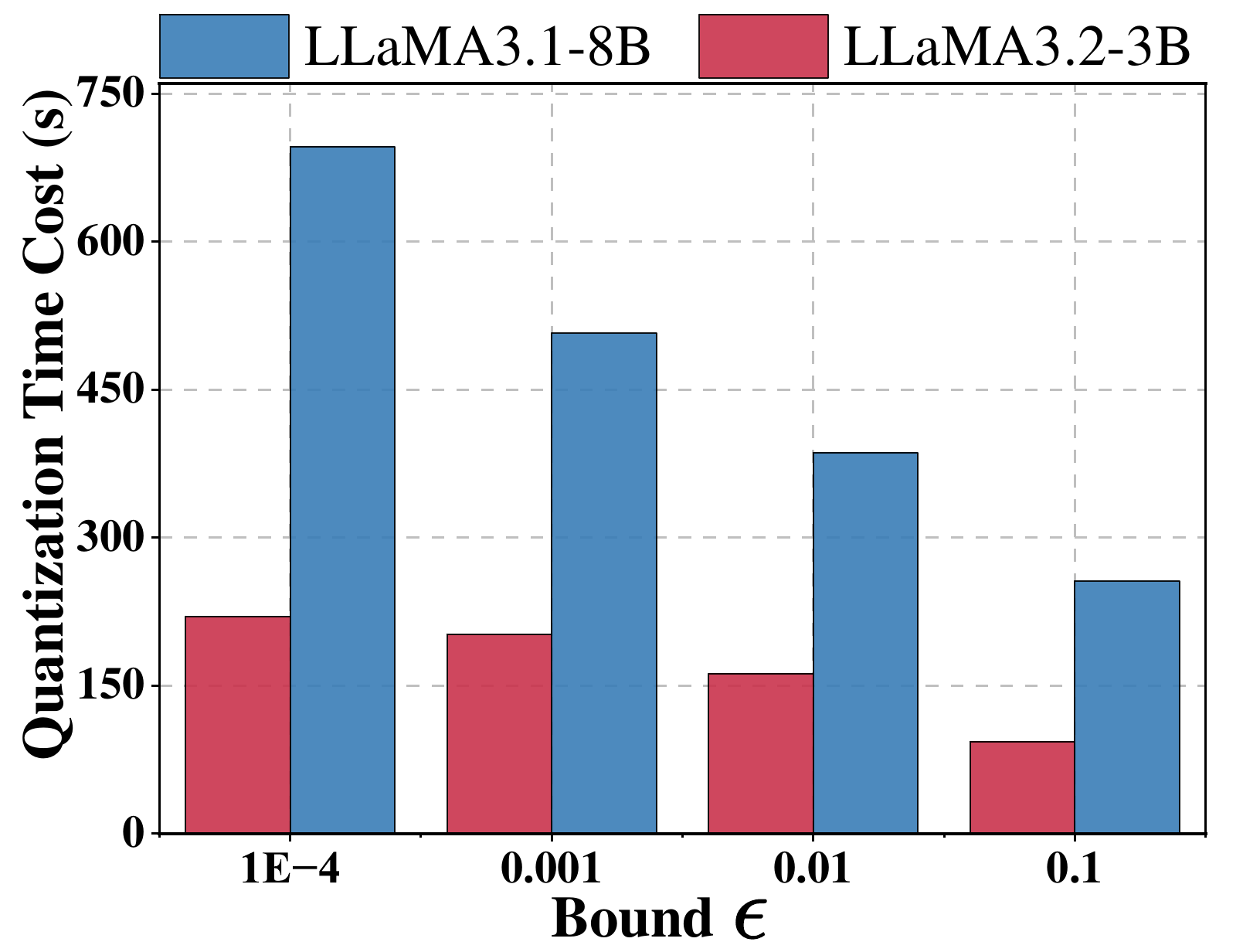}} 
		\subfigure[]{
			\includegraphics[width=0.45\linewidth]{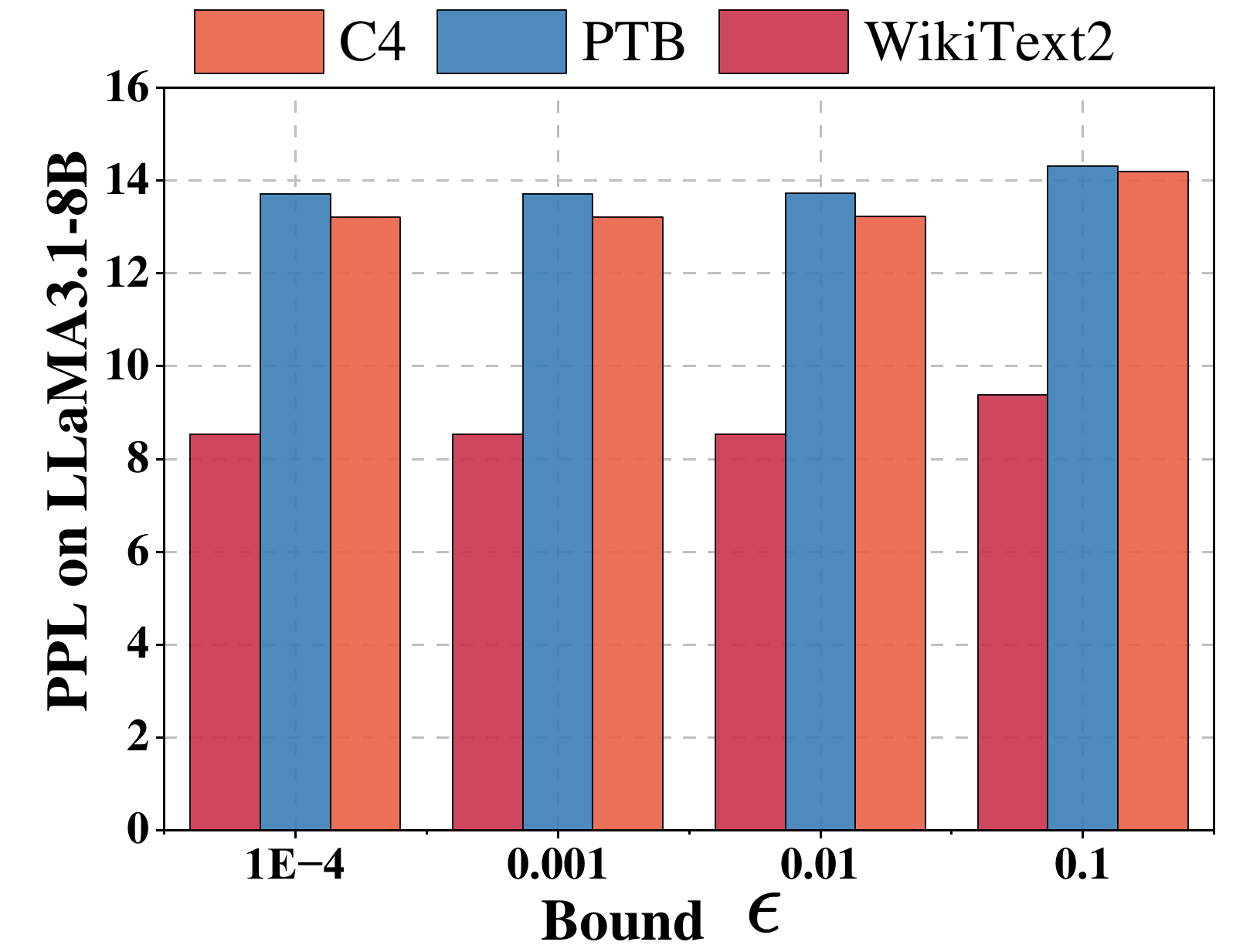}}  \\
		\subfigure[]{
			\includegraphics[width=0.45\linewidth]{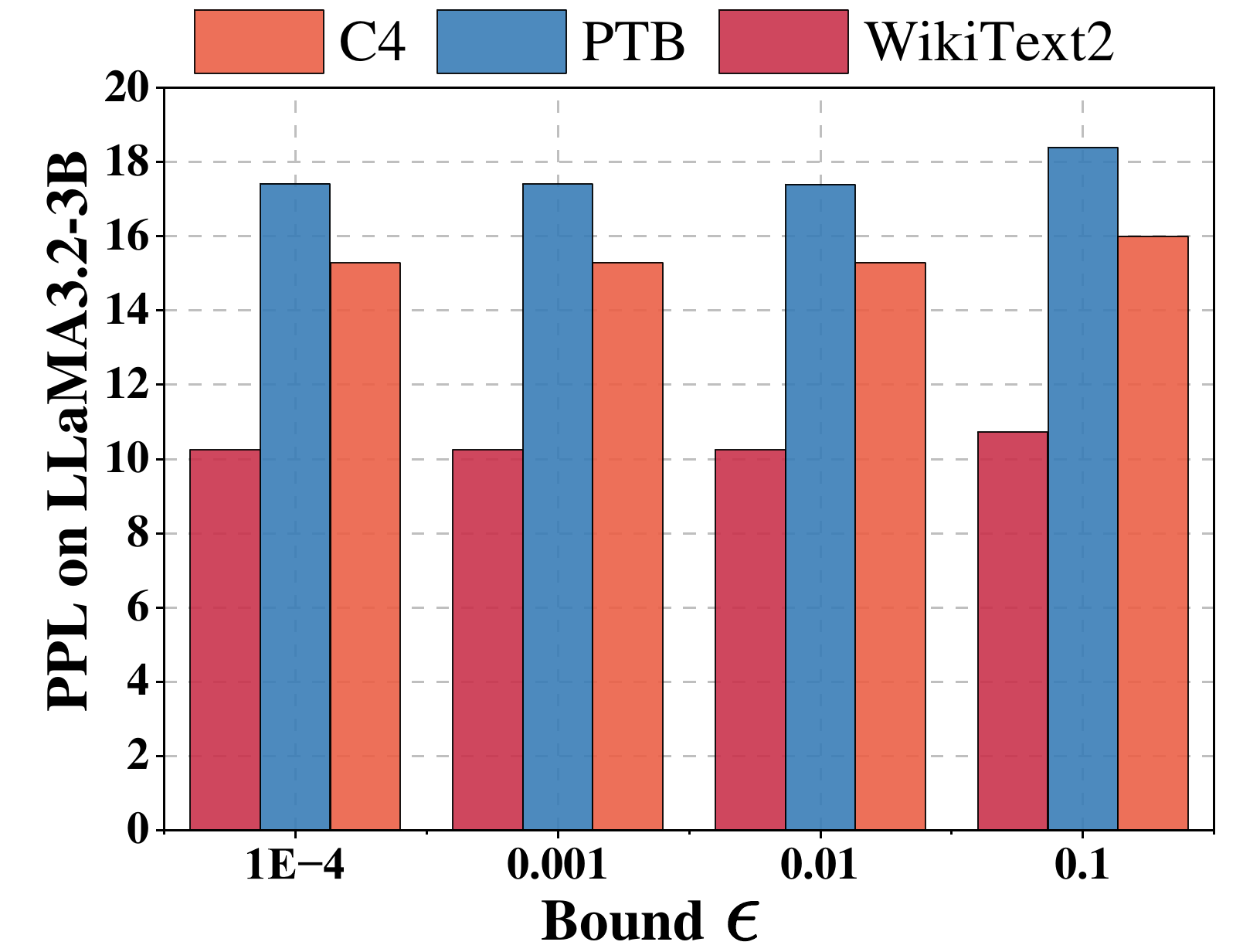}} 
		\subfigure[]{
			\includegraphics[width=0.45\linewidth]{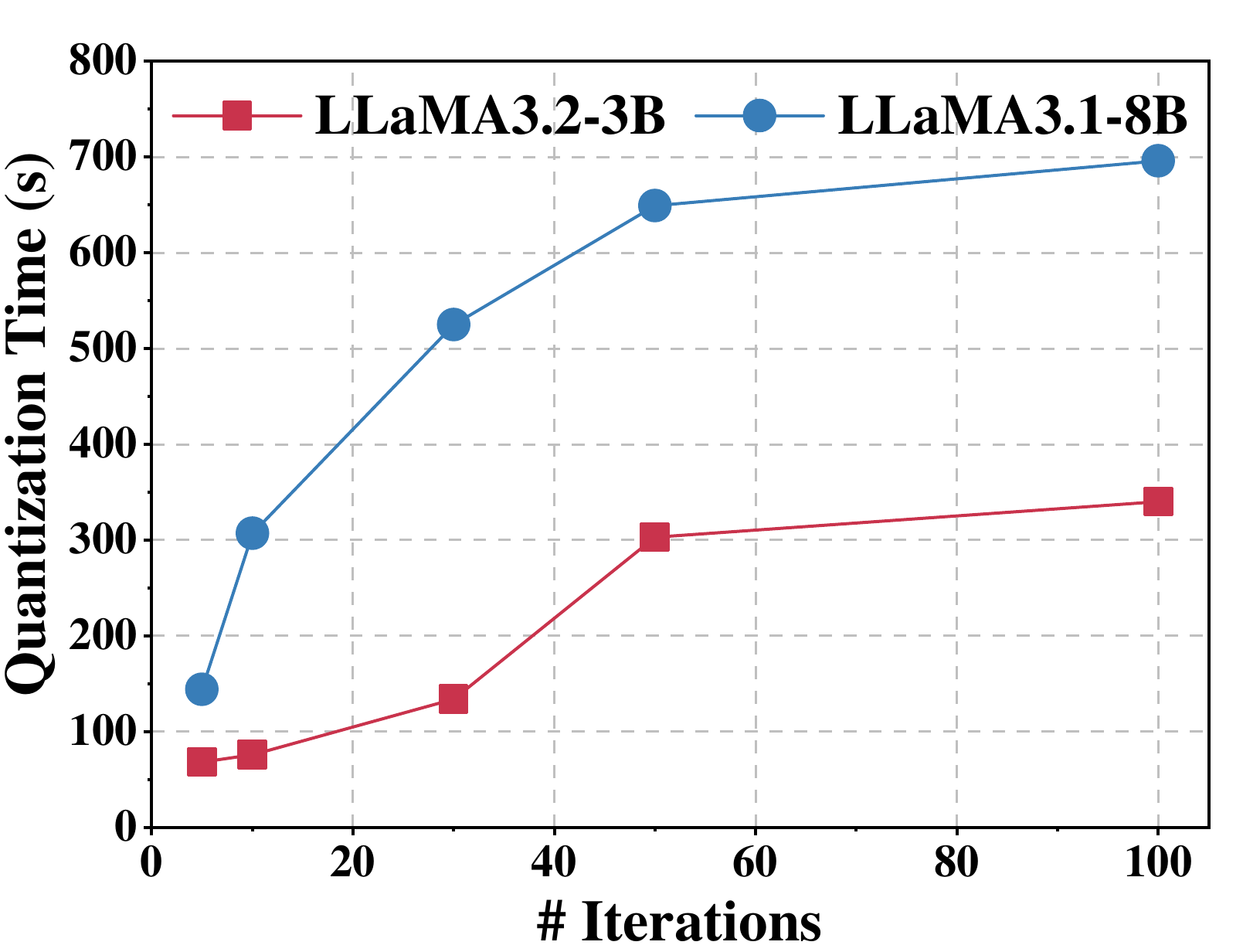}}  \\
		\subfigure[]{
			\includegraphics[width=0.45\linewidth]{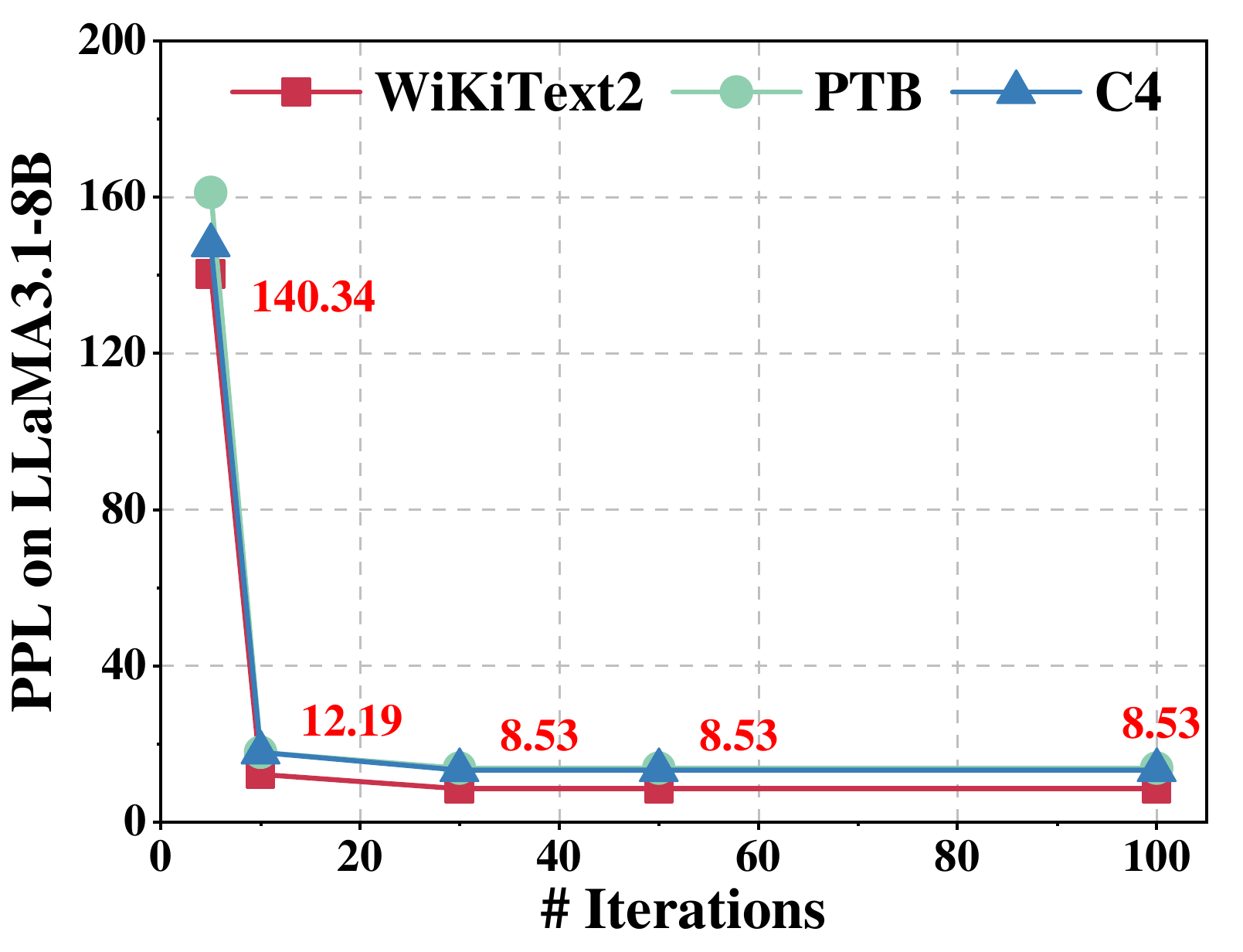}} 
		\subfigure[]{
			\includegraphics[width=0.45\linewidth]{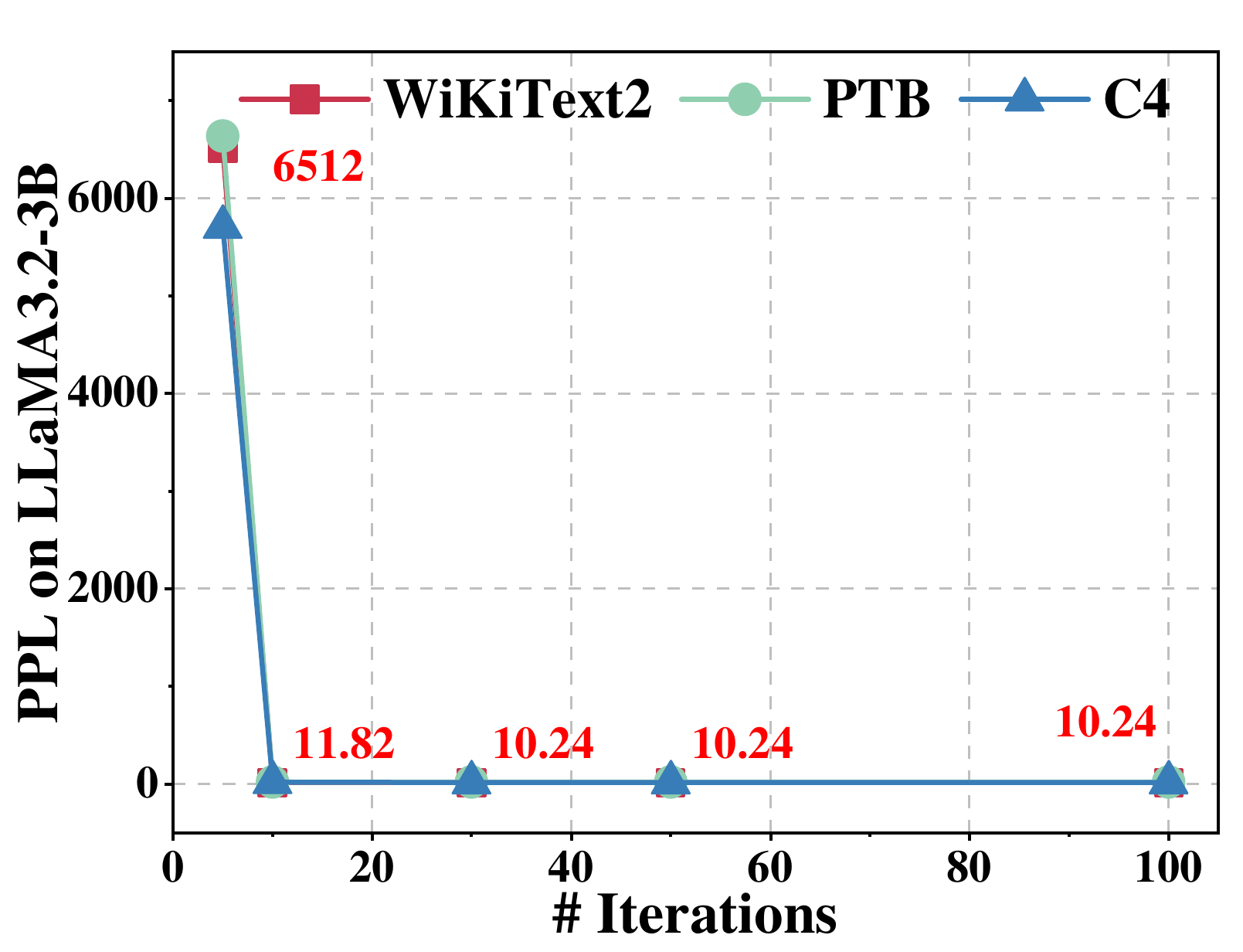}}
	\end{tabular}
    \vspace{-1em}
	\caption{(a)-(c).~Effect of progressive search iterations on quantization time (left sub-figure) and perplexity (PPL) (middle and right sub-figures) for LLaMA3.1-8B and LLaMA3.2-3B models. (d)-(f).~Trade-off between tolerance bounds ($\epsilon$), quantization time (left sub-figure), and model perplexity (PPL) (middle and right sub-figures) for LLaMA3.1-8B and LLaMA3.2-3B architectures.}
	\label{fig_bound}
    \vspace{-0.5em}
\end{figure}

\input{tables/table_add_revised}

\paragraph{Condition Numbers.}
{Table~\ref{tab:ppl} summarizes the validation perplexities obtained when sweeping a regularization coefficient (in log-space) across two model scales: LLaMA-3.1 8B and LLaMA-3.2 3B. All PPLs are reported on WikiText-2, PTB, and the C4 validation split. Across corpora, both models exhibit monotonically decreasing perplexity (illustrates robustness) as the condition numbers increases from $10^0$ to $10^2$, after which the metric saturates—indicating a regime where further rise the condition bounds offers negligible improvement.}

\paragraph{Progressive Search Iterations.}  
Fig. \ref{fig_bound}(a)-(c) identify a critical 30-iteration threshold that optimally balances perplexity and efficiency across model scales. Within this range, both test models achieve remarkable convergence.
% : LLaMA3.1-8B reduces WikiText2 perplexity by 94\%, while LLaMA3.2-3B recovers from catastrophic initialization by effectively adapting trit-plane coefficients to weight distributions. 
Smaller models converge faster due to simpler weight structures, while larger models leverage their expressivity for steeper initial gains. Beyond 30 iterations, perplexity stabilizes while quantization time increases linearly, indicating diminishing returns from additional refinement. 
% This confirms our theoretical guarantee of monotonic convergence and suggests limiting iterations $\leq$ 50 in practical compression process to achieve near-optimal accuracy.

\paragraph{Bound of Tolerance.}  
Fig. \ref{fig_bound}(d)-(f) reveal the crucial trade-off between tolerance \(\epsilon\) and quantization performance. Tighter \(\epsilon\) values improve perplexity but at significant computational cost: LLaMA3.1-8B achieves 9.1\% perplexity reduction with a 172\% runtime increase, while LLaMA3.2-3B shows 4.5\% improvement with 137\% longer runtime. 
% The inflection point at \(\epsilon=10^{-2}\) marks where returns diminish further tightening yields marginal improvements while significantly increasing computation time. 
Smaller models show better stability at low \(\epsilon\), while larger models require stricter tolerance to capture complex weight relationships. Our experiments identify \(\epsilon \in [10^{-3}, 10^{-4}]\) as the optimal range for balancing accuracy and efficiency, particularly important for resource-constrained deployment scenarios.

\section{Inference Efficiency}

\paragraph{Hardware Efficiency.}  
The PTQTP framework leverages ternary operations \(\{-1, 0, 1\}\) to achieve hardware-efficient computations. For a ternary element \(c_m \in \{-1, 0, 1\}\) and scaling coefficient $\alpha$, the multiplication operation with \(c_m\) in dequantization or dequantization-free forward can be implemented as an addition,  eliminating floating-point multipliers and replacing them with sign flips, identity mappings, or zeroing operations. This reduces the arithmetic intensity to \(\mathcal{O}(1)\) per element in the matrices. The far lower multiplication and the inherent sparsity in PTQTP than other low-bit methods and the decomposition into two trit-planes \(T^{(1)}, T^{(2)}\), allowing parallelized additive superposition of dual trit-planes, where each plane can be processed independently on parallelized architectures, exploiting data-level parallelism and minimizing latency in inference acceleration. This bandwidth reduction is critical for latency-bound applications, as memory access often dominates inference time in modern hardware architectures, and the final end-to-end experiments also verify the efficiency of PTQTP.

\paragraph{End-to-End Speed.}
To fully exploit the hardware-friendly properties of ternary weight quantization in PTQTP, we design a ternary matrix multiplication CUDA kernel based on LUT-GEMM \cite{Park2022LUTGEMMQM}. Since the weight values are restricted to \{-1,0,1\}, all possible dot products between activation vectors and low-bit weight patterns can be precomputed and stored in lookup tables (LUTs) in the shared memory. Notably, these dot products involve only addition operations. As a result, subsequent matrix multiplications can be performed via direct LUT lookups without any dequantization, and the computational overhead is limited to a one-time additive cost during LUT construction and shared memory read.

In our experiments, we evaluate the end-to-end generation throughput of PTQTP and other baseline methods (AQLM \cite{Egiazarian2024ExtremeCO} and QuIP\#) on a single 24GB RTX 3090 GPU. The input prompt length is set to 512 tokens, the generated sequence length is 512 tokens, and the batch size is 1. The results are summarized in Table \ref{tab:e2e}. For LLaMA-2 7B inference, PTQTP achieves a 4.63× speedup compared to the FP16 baseline. For LLaMA-2 13B, PTQTP delivers 3.57× and 1.53× speedups over AQLM and QuIP\#, respectively.

% Requires: \usepackage{booktabs}
\begin{table}[t]
    \centering
    % \small
    \caption{The end-to-end generation speed (tokens/s) comparison. *: b1.58-Dual. }
    \label{tab:e2e}
    \begin{tabular}{lccccc}
        \toprule
         & \small{FP16} & \small{AQLM-b2} & \small{QuIP\#-b2} & \small{PTQTP*} \\
        \midrule
        \small{L2-7B}  & \small{32.2} & \small{65.7} & \small{99.04} & \small{149.35} \\
        \small{L2-13B} & \small{OOM}  & \small{26.3} & \small{61.32} & \small{93.98}  \\
        \bottomrule
    \end{tabular}
\end{table}

\section{Conclusion}

We introduced PTQTP, a structured PTQ framework that achieves ternary quantization through systematic trit-plane decomposition without retraining or fine-tuning. Extensive experiments demonstrate that PTQTP outperforms existing low-bit PTQ methods, and approaches or even surpasses the accuracy of 4-bit PTQ and 1.58-bit QAT techniques. Remarkably, PTQTP preserves mathematical reasoning and coding capabilities that catastrophically degrade or collapse in other extremely low-bit PTQ schemes, challenging the conventional wisdom that such tasks inherently require higher precision. In addition, PTQTP's model-agnostic design ensures robust deployment across diverse architectures from 1B to 70B models while providing a impressive end-to-end speed gains. By addressing the core challenges of expressiveness, stability, scalability, and hardware efficiency, PTQTP sets a new yardstick for compressing high-performance LLMs for resource-constrained platforms.

\subsubsection*{Limitations}
PTQTP's memory layout can be further optimized through bit-packing, where 8 ternary elements are stored in a single byte, improving cache locality and reducing cache misses by 20\%-30\% compared to non-packed formats. In addition, PTQTP's multiplication-free operations align perfectly with hardware accelerators designed for binary/ternary neural networks, such as customized ASICs. These architectures can execute sign flips and additions in single clock cycles, achieving peak throughput with minimal energy consumption. For instance, ternary operations can leverage bitwise XOR for sign inversion, enabling sub-nanosecond latency per operation on specialized hardware. However, currently,  technical support is still not widespread, and PTQTP's storage and technical computing still rely on modern computing devices, which limits the performance of PTQTP to some extent. Future works can focus on software-hardware co-design and hardware framework design for 1.58-bit inference acceleration.

% \textbf{Generalization Optimization.} Although we have tested on various benchmarks to showcase the advancement and robustness of PTQTP as much as possible, due to time constraints, we will challenge more difficult benchmarks in future work, which has been lacking in previous PTQ-related work. We may further enhance the performance of PTQTP through methods similar to DPO or SFT, enabling it to provide stable support for all future quantization-required scenarios.
\subsubsection*{LLM Usage Disclosure}
This paper is primarily the work of the human authors, but we also made use of several advanced LLMs, including ChatGPT-5 and Gemini 3. They were employed to assist with code development and debugging, support result analysis, and help with formatting and language polishing. We acknowledge the contributions of these LLMs, while fully recognizing their limitations, and take full responsibility for all content presented under our names. We did not include hidden prompt-injection text in the submission, and all external data and code comply with their respective licenses.

% \section*{Acknowledgments}


%%%%%%%%%%%%%%%%%%%%%%%%%%%%%%%%%%%%%%%%%%%%
\bibliography{custom}
%%%%%%%%%%%%%%%%%%%%%%%%%%%%%%%%%%%%%%%%%%%%

\appendix

\section{Related Works}
% \label{sec:background}
\paragraph{Recent Progress in PTQ for LLMs.}
PTQ has emerged as a promising approach for compressing LLMs without the computational overhead of retraining. GPTQ \citep{frantar-gptq} pioneered efficient 4-bit quantization through layer-wise optimization. AWQ \citep{jlin2024AWQ} further improved this by incorporating activation-aware scaling, demonstrating that the consideration of activations during quantization leads to better performance. Recent advancements like QuIP\# \citep{quip1}, AQLM \citep{aqlm}, and \citep{huang2024slim,chee2023quip,zhao2025ptq1.61} have pushed the boundaries of PTQ by utilizing vector quantization or learnable equivalent transformations to handle outliers. While these methods achieve impressive accuracy, they often introduce non-trivial decoding overhead (e.g., codebook lookups) or require complex calibration optimization. LieQ~\citep{lieq} attempts to do the structured and explainable quantization using layer-wise mixed precision scheme,achieving extremely low-bit PTQ with well-reasoned and hardware-friendly properties. In addition, other works have pushed towards lower bit width PTQ. PBLLM \citep{shang2023pbllm} and  \citep{huang2024billm,li2025arbllm} achieved 1-bit quantization by identifying and preserving structurally salient weights while applying different quantization strategies to different weight groups. However, these methods rely on complex, unstructured weight categorizations that complicate hardware implementation. 

\paragraph{Ternary Quantization.}
The transition from binary to ternary quantization represents a leap of model expressiveness while still preserving multiplier-less arithmetic. BitNet's \citep{BitNet} extension to 1.58-bit (ternary) weights \citep{ma2024bitnet1.58} demonstrated superior performance compared to binary models, suggesting the value of the additional state in the ternary representation. However. existing ternary quantization methods mainly (if not all) rely on QAT approaches (such as BitNet), requiring extensive pretraining and substantial computational resources. In addition, previous methods use unstructured weight quantization to compensate for the degenerate representation ability. Unlike these works that optimize progressive reconstruction via retrained context models, PTQTP is a post-training, structured and model-agnostic framework that decomposes LLM weights into uniform and collaborative trit-planes with adaptive ridge-scaled coefficients to yield 1.58-bit quantization without retraining or fine-tuning. 

\paragraph{Hardware Efficiency.} A key advantage of binary and ternary quantization lies in their suitability for FPGA or ASIC deployment (e.g., FlightLLM \citep{zeng2024flightllm}, DFX \citep{DFX}), where multiplier-less feature–weight products can be efficiently implemented in hardware. Specifically, binary operations can be realized with XNOR gates and bit-counting, while ternary operations leverage simple adders, preserving computational efficiency while offering greater expressivity. Therefore, PTQTP focuses on the specific niche of \textit{multiplication-free} inference via structured ternary decomposition, prioritizing hardware efficiency on edge devices where adders are significantly cheaper than multipliers.

% \section*{Appendix}
\section{More Perspective on Efficiency Analysis } 

\begin{figure}[tbp]
  \centering 
  \includegraphics[width=0.45\textwidth]{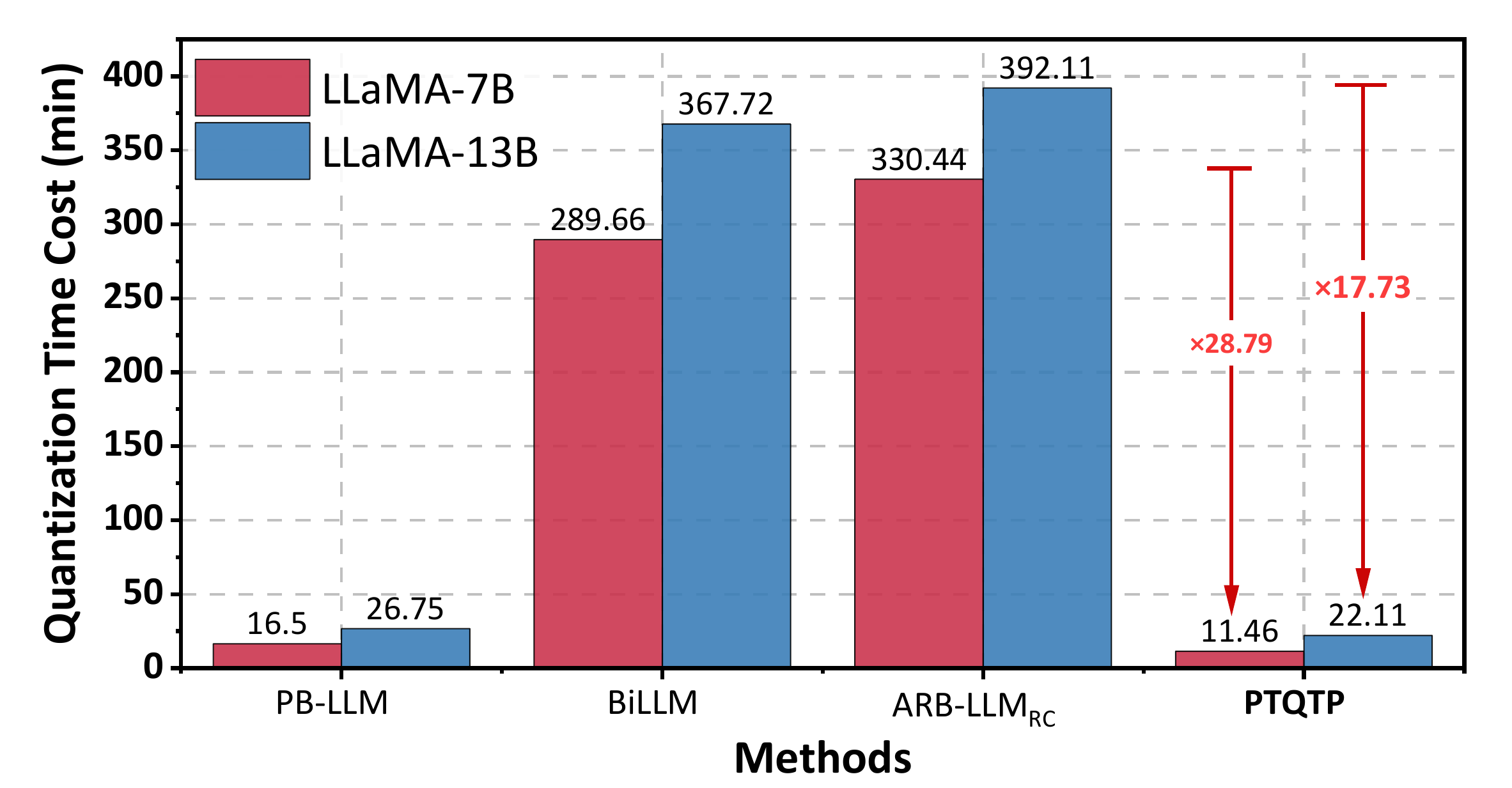} 
\caption{Runtime performance evaluation. PTQTP outperforms existing extremely low-bit quantization methods while achieving 17.73$\times$ -- 28.79$\times$ speedup over ARB-LLM\textsubscript{RC}.}
  \label{appendix_runtime} 
\end{figure}

\paragraph{Computational Complexity.} For row \(i\), the normal equation matrix \(A_i = (S_i)^T S_i + \lambda_iI_2\) in PTQTP is \(2 \times 2\), with inverse computed in constant time \(\mathcal{O}(1)\) using the formula:  
   \begin{equation}
   A_i^{-1} = \frac{1}{\det(A_i)} \begin{bmatrix} A_i(2,2) & -A_i{(1,2)} \\ -A_i(2,1) & A_i(1,1) \end{bmatrix}
   \end{equation}
Where \(\det(A_i) = A_i(1,1)A_i(2,2) - A_i(1,2)^2\). The vector \(b_i = S_i^T W_i^T\) requires \(2d\) operations (dot products for each column of \(S_i\)), leading to \(\mathcal{O}(d)\) per row. Adding up to \(n\) rows, this step is \(\mathcal{O}(nd)\) per iteration.  In addition, for each element \((i,j)\) in $W$, evaluating $(T^{(1)}, T^{(2)}) \in \{-1, 0, 1\}^2$ involves \(9\) arithmetic operations per element, resulting in \(\mathcal{O}(1)\) complexity per element. For \(d\) elements per row, this is \(\mathcal{O}(nd)\) in \(n\) rows per iteration. Therefore,  with \(T_{\text{max}}\) iterations, the total computational complexity is $\mathcal{O}(T_{\text{max}} \cdot nd)$.
% \begin{equation}
% \mathcal{O}\left(T_{\text{max}} \cdot (nd + nd)\right) = \mathcal{O}\left(T_{\text{max}} \cdot nd\right).
% \end{equation}
 
\paragraph{Time Complexity.} For each iteration, the time complexity can be presented as $\mathcal{O}(nd)$.
    % \begin{equation}
    %      \underbrace{\mathcal{O}(nd)}_{\substack{\text{Ridge Regression} \\ \text{solve for } \alpha}} + 
    %   \underbrace{\mathcal{O}(nd)}_{\substack{\text{Trit-Plane} \\ \text{element-wise search}}} = \mathcal{O}(nd);
    %    \underbrace{ \mathcal{O}(n)}_{\substack{\text{Convergence} \\ \text{Check}}} \ll \mathcal{O}(nd)
    % \end{equation}
With empirical convergence in \(T_{\text{max}} \leq 50\) iterations, the total time complexity is $\mathcal{O}\left(T_{\text{max}} \cdot nd\right)$, exhibiting linear scalability with the dimension of weight matrix  \(n, d\), critical for deployment on LLMs with billions of parameters. 

\(\mathcal{O}(T_{\text{max}} \cdot nd)\) complexity is optimal for low-bit quantization when considering both approximation quality and computational efficiency. Compared to existing PTQ methods, \textbf{\textit{PTQTP achieves lower per-iteration complexity due to its closed-form solutions and fixed-dimensional ridge regression.} }

% \subsection{Ternary vs Binary PTQ}  
\paragraph{Run Time Consumption.}
The experiment results in Fig. \ref{fig_bound} show that PTQTP can stably converges less than 50 iterations with different model structures, resulting in 28.79 \(\times\) speedup during quantization on modern GPUs, which are shown in Fig. \ref{appendix_runtime}(a). In other words, PTQTP can quickly find the optimal trit-planes to achieve model compression. Compared to existing extremely low-bit schemes, it significantly reduces runtime overhead, which we believe is mainly due to the absence of additional outlier handling and protection operations. Furthermore, PTQTP also boasts the hardware efficiency advantages of binary compression, and its performance is stronger and more stable. More performance comparison are shown in Appendix~\ref{app_more_results}.

\paragraph{Memory Saving.}
For $W \in \mathbb{R}^{n\times d}$, group size $k$, the memory of $\hat{W}$ after standard quantization of bits $m$ is
\begin{align}
\mathcal{M} =&\ 
\underbrace{n\times d \times m}_{B} + \underbrace{[d/k]}_{\text{multiple groups}} \times \underbrace{n\times 16}_{\text{row-wise FP16 $\alpha$}}
\end{align}

The memory required by BiLLM can be formulated as follows, where $c$ is the number of salient columns for $W$.
\begin{align}\label{eq:BiLLM_mem}
\mathcal{M}_{\text{BiLLM}} =&\underbrace{n\times d}_{\text{group bitmap}} + \underbrace{d}_{\text{salient column bitmap}}\notag\\
&+ \underbrace{2 \times n \times c + [d/k] \times 3n \times 16}_{\text{second-order binarization}} \notag\\
&+ \underbrace{ n \times (d - c) + \overbrace{[d/k] \times 2n\times 32}^{\text{2 groups}}}_{\text{first-order binarization}}  
\end{align}

Previous work~\citep{li2025arbllm} derived the total memory occupation ARB-RC and ARB-RC + CGB (grouped column bitmap), which can be formulated as:
\begin{align}\label{eq:ARB_mem}
&\mathcal{M}_{\text{ARB-RC}} =\underbrace{n \times d}_{\text{group bitmap}} + \underbrace{d}_{\text{salient column bitmap}}\notag\\
&+ \underbrace{2 \times n \times c + \left( [d/k]] \times 2n + 2c \right) \times 16}_{\text{second-order binarization}} \notag\\
&+ \underbrace{n \times (d - c) + \overbrace{\left( [d/k] \times n + (d - c) \right) \times 16 \times 2}^{\text{2 groups}}}_{\text{first-order binarization}}
\end{align}

\begin{align}\label{eq:ARBRC_mem}
&\mathcal{M}_{\text{ARB-RC+CGB}}=\underbrace{n \times d}_{\text{group bitmap}} + \underbrace{d}_{\text{salient column bitmap}}\notag\\
& + \ \underbrace{2 \times n \times c + \left( [d/k] \times 2n + 2c \right) \times 16 \times 2}_{\text{second-order binarization}} \notag\\
& + \underbrace{n \times (d - c) + \overbrace{\left( [d/k] \times n + (d - c) \right) \times 16 \times 2}^{\text{2 groups}}}_{\text{first-order binarization}}
\end{align}

In PTQTP, each trit-plane containing 3 states has to be stored as a 2bit datatype due to the hardware constraint. The total memory of PTQTP is
\begin{align}
\mathcal{M}_{\text{PTQTP}} =\ 
&\underbrace{ 2 \times n\times d \times 2}_{\text{2 Trit-Plane}} + \underbrace{[d/k]\times 2n\times 16}_{\text{row-wise FP16 $\alpha$}}
\end{align}

Figure~\ref{appendix_runtime}(b) demonstrates the estimated memory demand of PTQTP with other binary quantization methods, derived from the above formulas. The proposed PTQTP slightly increased the memory consumption to other binary quantization approaches. This is a trade-off between storage and representational capacity. However, methods such as BiLLM and ARB-LLM\textsubscript{RC} explicitly divide columns into first-order and second-order groups based on their saliency (as shown in Eqs.\ref{eq:BiLLM_mem}, \ref{eq:ARB_mem}, \ref{eq:ARBRC_mem}), assigning more bit planes and FP16 parameters to the more salient second-order columns. As a result, \textit{\textbf{although PTQTP uses trit-planes to represent quantized weights, it does not incur significant memory overhead compared to binary-based methods}}.

\section{Implementation Details About PTQTP}
% \label{appalg}
\paragraph{Progressive and Adaptive Regularization.}
Algorithm~\ref{alg_trit_decomp} illustrates more details about the process of PTQTP. In the beginning, the trit-planes \(T^{(k)}\) are initialized using the sign function of \(W\), with zero entries replaced by 1 to ensure valid ternary values (later expanded to include 0 through optimization); scaling coefficients \(\alpha\) are initialized as uniform vectors \([1, 1]\) replicated across all rows; and the regularization parameter \(\lambda\) starts at a small value \(10^{-8}\) to promote numerical stability.

\input{algs/arg}

In each iteration, the algorithm first updates the continuous scaling coefficients \(\alpha\) row by row. 
If the condition number \(\kappa_{i,approx} \geq 10^{6}\), indicating ill-conditioning, the regularization parameter \(\lambda_i\) is adaptively increased  \(\leq\lambda_{max} = 1.0\) to stabilize the solution. This adaptive regularization mitigates the sensitivity to small input perturbations. The scaling coefficients are then solved via closed-form ridge regression, ensuring efficient computation without iterative solvers. Next, the algorithm optimizes the discrete trit-planes. It generates all 9 possible ternary value combinations \(\mathcal{C} = \{-1, 0, 1\}^2\) for the pair \((T^{(1)}, T^{(2)})\), and for each matrix element \((i,j)\), computes the approximation error for each combination. The combination \(\mathbf{c}_{m^*}\) that minimizes the squared error $e_m$ is selected to update \(T^{(1)}_{ij}\) and \(T^{(2)}_{ij}\), effectively conducting a local exhaustive search to find the best ternary representation for each weight entry. Finally, convergence is checked by monitoring the Frobenius norm difference between consecutive scaling coefficient updates. If \(\|\alpha_{(t)} - \alpha_{(t-1)}\|_F < \epsilon\), the algorithm terminates early, balancing optimization quality with computational efficiency. 

\paragraph{Regularized Initialization.} 
We adopt a progressive initialization strategy to kick-start the optimization process. The first step of our initialization focuses on approximating the weight matrix \(\tilde W_i\) group by group. We start by initializing the trit-planes \(\tilde T^{(1)}_{i,(0)}\) \(\tilde T^{(2)}_{i,(0)}\)using the sign function of \(\tilde W_i\):
\begin{align}
    \tilde T^{(k)}_{i,(0)} & = \text{sign}(\tilde W_i)
\end{align}

After obtaining the group-wise initialized trit-planes \( \tilde T^{(1)}_{i,(0)} \) and \( \tilde T^{(2)}_{i,(0)} \), we construct the matrix \( \tilde S_{i,(0)} \) where each row contains elements from \( \tilde T^{(1)}_{i,(0)} \) and \( \tilde T^{(2)}_{i,(0)} \). The initial scaling coefficients \( \tilde \alpha_{i,(0)}  \) are obtained by solving:

\begin{equation}
    \tilde \alpha_{i,(0)} = \arg\min_{\alpha} \left( \|  \tilde W_i - \tilde S_{i,(0)}\theta  \|_F^2 + \lambda \| \theta \|_F^2 \right)
\end{equation}

The closed-form solution with dimension-compatible regularization becomes:
\begin{equation}
    \tilde \alpha_{i,(0)} = \left( \tilde S_{i,(0)}^T \tilde S_{i,(0)} + \lambda I \right)^{-1} \tilde S_{i,(0)}^T \tilde W_i^T
\end{equation}

The coefficient bound should be modified as follows, where $\sigma_{\max}(\cdot)$ denotes the maximum singular value.
\begin{equation}
    \| \tilde \alpha_{i,(0)} \|_2 \leq \frac{\sigma_{\max}(\tilde S_{i,(0)})}{\sigma_{\min}^2(\tilde S_{i,(0)}) + \lambda} \| \tilde W_i^T \|_2
\end{equation}

\section{More Performance Illustration}
\label{app_more_results}
\paragraph{Full experiments results of PTQTP.}
In this supplementary chapter, we listed all our test results. Table \ref{tab_qwen3-FP16-PTQTP} displays the results of the benchmark experiment with a certain level of difficulty, showing the quantization stability of our PTQTP for the Qwen3-14B quantized model.  Table \ref{tab_compare_qat_more} illustrated the specific accuracy comparison about PTQTP on 6 zero-shot reasoning tasks. 
\input{tables/Downstream_qwen3}

\paragraph{Performance vs Binary PTQ.}
We list all of our experiment results here for reference. We further compared PTQTP performance on different datasets with various model size and backbones, the results are illustrated in Table \ref{tab_llama_ppl} and Table \ref{tab_quanT^14b}. 
According to these results, we find specialized binary schemes (PBLLM, BiLLM) show catastrophic failure on Math-500 (0\%) and GSM8K ($<$2\%). Even ARB-LLM\textsubscript{RC}, despite architectural modifications, achieves only 1.80\% and 32.22\% on these benchmarks—significantly below baseline. \textit{\textbf{This evidence confirms that mathematical reasoning is uniquely vulnerable to precision loss in conventional PTQ approaches}}, exhibiting 19$\times$ greater performance degradation compared to linguistic tasks, with non-negligible impacts (12.7\% mean reduction) on general reasoning benchmarks. These findings suggest that architectural modifications alone—such as salient weight protection—cannot effectively preserve mathematical reasoning capabilities at ultra-low precision. In contrast, PTQTP-b1.58 achieves 82.40\% on Math-500 and 85.44\% on GSM8K—less than 5\% degradation from the baseline. This preservation of mathematical reasoning at ultra-low bit-widths suggests PTQTP effectively decouples numerical precision requirements from model parameterization. 

% \section{Broader Impacts}
% Our research fundamentally reconstructs the computational paradigm of artificial intelligence (AI) edge systems by eliminating the vast majority of multiplication operations. This breakthrough enables novel hardware innovations, particularly in the development of pure additive-operation-based neuromorphic chip designs, significantly enhancing both thermal efficiency and computational stability in mobile devices. The advancement dramatically lowers the barrier to AI deployment, making serverless architectures feasible even in energy-constrained regions. 
% At the mathematical framework level, the established non-multiplicative tensor algebra system opens new research directions in biologically inspired dendritic computing. Although challenges remain in addressing the numerical stability of additive accumulations and the co-design of next-generation AI accelerators, this computational paradigm shift has already reshaped the energy-performance tradeoff curve, providing new pathways for existing AI infrastructure development.
\clearpage

\input{tables/benchmark_all}
\input{tables/llamappl}
\input{tables/benchmark_qwen3_14b}

\end{document}

%% file: algs/app_group_algro.tex
\begin{algorithm}[tbp]
\caption{Approximation Process of PTQTP}
\label{alg_group}
\begin{algorithmic}[1]
\Require Weight matrix $W$, group size $G$, max iterations $T_{\max}$, tolerance $\epsilon$
\Ensure Quantized weight approximation $\hat{W}$

% \State \textbf{Group Initialization:}
\State Divide $W$ into groups $\tilde W_i \in \mathbb{R}^{G}$ 
\State Initialize $\tilde T^{(k)}_{i,(0)} \gets \text{sign}(\tilde W_i)$, $\tilde\alpha^{(k)}_{i,(0)} \gets [1, 1]$ 

\For{$t \gets 1$ to $T_{\max}$}
    % \State \textbf{Group Scaling Update:}
    \For{each row $i$} row-wise decomposition
        \State $\tilde S_i \gets [(\tilde T^{(1)}_{i,(t-1)})^T \quad (\tilde T^{(2)}_{i,(t-1)})^T]$ 
        \State $\tilde\alpha^{(k)}_{i,(t)} \gets \text{RidgeRegression}(\tilde S_i, \tilde W_i, \lambda_i)$
    \EndFor
    
    % \State \textbf{Trit-Plane Optimization:}
    \For{each row $i$} 
        \State $\tilde T^{(1)}_{i,(t)}, \tilde T^{(2)}_{i,(t)} \gets \arg\min_{c_m^{(1)},c_m^{(2)}} \|\tilde W_i - \tilde\alpha^{(1)}_{i,(t)}c_m^{(1)} - \tilde\alpha^{(2)}_{i,(t)}c_m^{(2)}\|^2_F$
    \EndFor
    % \Comment{Spatial position parallel optimization}
    
    % \State \textbf{Convergence Check:}
    \If{$\max\limits_i \|\alpha^{(k)}_{i,(t)} - \alpha^{(k)}_{i,(t-1)}\|_F < \epsilon$} \textbf{break}
    \EndIf
\EndFor

% \State \textbf{Combine:} $\hat{W} \gets \bigoplus\limits_g \sum_{k=1}^2 \text{diag}(\alpha^{(k)}_i) T^{(k)}_i$ \Comment{Group-wise concatenation}
\end{algorithmic}
\end{algorithm}

%% file: tables/main_results.tex
\begin{table*}[ht]
\centering
\small
\setlength{\tabcolsep}{2pt}
\caption{Perplexity ($\downarrow$) comparison across SOTA model families on WikiText2 and C4 datasets (\textit{group size=128}). Fine-Tuning means the further optimization step in the quantization method; Salience detect means there are extra search and protect salience weight step in the method. L2: LLaMA2; L3: LLaMA3; Q3: Qwen3; N/A: Model size variant not available; D$^\dagger$: Dual; \textbf{Bold}: best result; \underline{underlined}: second-best.}
\label{tab_qwen}
\begin{tabular}{l@{\hspace{4pt}} c@{\hspace{1pt}} c@{\hspace{1pt}} c@{\hspace{2pt}} c@{\hspace{2pt}} c@{\hspace{2pt}} c@{\hspace{2pt}} c@{\hspace{2pt}} c@{\hspace{2pt}} c@{\hspace{2pt}} c@{\hspace{2pt}} c@{\hspace{4pt}} c@{\hspace{2pt}} c@{\hspace{2pt}} c}
\toprule
&&& \multicolumn{9}{c}{WikiText2} & \multicolumn{3}{c}{C4} \\
\cmidrule(lr){4-12}\cmidrule(lr){13-15}
Method & \begin{tabular}{c} Fine\\Tuning \end{tabular}  & \begin{tabular}{c} Salience \\ Detect \end{tabular} & L2-7 & L3-1 & L3-3 & L3-8 & L3-70 & Q3-1.7 & Q3-4 & Q3-8 & Q3-32 & L2-7 & L3-8 & L3-70 \\
\midrule
FP16     & N/A & N/A & 5.47  & 9.75 & 7.80  & 6.23 & 2.81 & 16.70 & 13.64 & 9.71 & 8.64 & 7.26 & 9.54 & 7.17  \\
\midrule
% AWQ-b4   & {\color{checkgreen}$\times$}  & {\color{checkgreen}$\times$}       & 5.61   & N/A & N/A  & 5.92 & 2.96 & 8.80 & 7.36 & 6.45 & N/A & 7.49 & 10.36 & 8.98 \\
QTIP-b4  & {\color{checkgreen}$\times$} & {\color{checkgreen}$\times$}      & \textbf{5.69}   & N/A & N/A  & \textbf{5.69} & \textbf{2.75} & \textbf{8.46} & \textbf{7.09} & \textbf{6.28} & N/A & \textbf{6.63}  & \textbf{7.22}  & \textbf{5.83} \\
GPTQ-b3  & {\color{checkgreen}$\times$}  & {\color{checkgreen}$\times$}       & 6.44   & \underline{18.99}  & \underline{16.34}  & 9.12   & 8.14   & 6.17e4 & 1.34e5 & 5.02e5 & \underline{37.10} & 7.95  & 17.68  & 10.04 \\
AWQ-b2   & {\color{checkgreen}$\times$}  & {\color{checkgreen}$\times$}       & 2.22e5 & 1.64e5 & 1.11e5 & 7.98e5 & 4.52e4 & 7.52e6 & 1.38e7 & 1.21e7 & N/A   & N/A  & N/A  & N/A \\
GPTQ-b2  & {\color{checkgreen}$\times$}  & {\color{checkgreen}$\times$}       & 52.22  & 4.97e3 & 2.42e3 & 717.54 & 18.79  & 655.10 & 5.92e5 & 7.77e4 & 38.40 & 35.27 & 394.74 & 122.55 \\
% PAROQ-b4    & {\color{crossred}$\checkmark$}  & {\color{checkgreen}$\times$}  & N/A    & N/A & N/A & N/A & N/A & N/A & N/A & N/A & N/A & N/A    & N/A & N/A & N/A & N/A \\
\midrule
OmniQuant-b2 & {\color{crossred}$\checkmark$} & {\color{checkgreen}$\times$}         & 11.06 & 771.05 & 538.64 & 2.08e3  & N/A & N/A & 5.30e4 & 7.31e4  & N/A & 15.02  & N/A      & N/A \\
QUIP\#-b2 & {\color{crossred}$\checkmark$} & {\color{crossred}$\checkmark$}    & 39.73 & N/A & N/A & 84.97 & N/A & N/A & N/A & N/A & N/A & 31.94  & 130.00 & N/A \\
% QTIP-b2     & {\color{crossred}$\checkmark$}  & {\color{crossred}$\checkmark$}   & N/A     & N/A & N/A & N/A     & N/A & N/A & N/A & N/A & N/A & N/A      & N/A      & N/A \\
% AQLM-b2     & {\color{crossred}$\checkmark$}  & {\color{crossred}$\checkmark$}   & N/A     & N/A & N/A & N/A     & N/A & N/A & N/A & N/A & N/A & N/A      & N/A      & N/A \\
SliM-LLM-b2  & {\color{crossred}$\checkmark$}  & {\color{crossred}$\checkmark$}  & 15.84 & 3.31e2 & 8.19e2 & 31.52 & N/A & 4.41e4 & 39.71  & N/A       & N/A & 84.92  & 390.02 & N/A \\
PBLLM-b1.7   & {\color{crossred}$\checkmark$}  & {\color{crossred}$\checkmark$}  & 66.41 & 3.87e3 & 1.01e4 & 1.89e3 & N/A & 9.27e5 & 4.68e3 & 1.80e3  & N/A & 80.69  & 104.15 & N/A \\
PTQ1.61-b1.61 & {\color{crossred}$\checkmark$}  & {\color{crossred}$\checkmark$} & 12.70 & N/A & N/A & 22.90 & N/A & N/A & N/A & N/A & N/A & N/A      & 33.82  & N/A \\
\textbf{PTQTP-b1.58-D$^\dagger$} & {\color{checkgreen}$\times$}   & {\color{checkgreen}$\times$}    & \underline{6.30}  & \textbf{17.15} & \textbf{10.24} & \underline{8.53} & \underline{7.76} & \underline{32.46} & \underline{18.25} & \underline{11.80} & \textbf{10.06} & \underline{6.76} & \underline{13.24} & \underline{12.64} \\

\bottomrule
\end{tabular}
\end{table*}

%% file: tables/New_table2_L2Q3.tex
\begin{table*}[t]
% \scriptsize
\small
\setlength{\tabcolsep}{4pt}
\centering
\caption{Zero-shot Reasoning Tasks Results on LLaMA2-7B and Qwen3-14B models (accuracy \%). D$^\dagger$: Dual; \textbf{Bold}: best result; \underline{underlined}: second-best. OBQA: OpenBookQA.}
\label{tab_zero_shot_results}
% \resizebox{\linewidth}{!}{%
\begin{tabular}{@{}l*{8}{c}@{}}
\toprule
\textbf{Method} &
\textbf{ARC-C} &
\textbf{ARC-E} &
\textbf{BoolQ} &
\textbf{HellaSwag} &
\textbf{OBQA} &
\textbf{PIQA} &
\textbf{WinoG} &
\textbf{Average} \\
\midrule
\multicolumn{9}{c}{\textbf{LLaMA2-7B}} \\
\midrule
FP16 & 43.09 & 75.46 & 79.36 & 57.11 & 33.2 & 78.13 & 69.38 & 62.25 \\
\midrule
GPTQ-b4 & \textbf{42.41} & \textbf{75.8} & \textbf{77.58} & \textbf{56.45} & \textbf{32.8} & \textbf{78.67} & \textbf{70.32} & \textbf{62.01} \\
% AWQ-b4 & 43.09 & 75.93 & 78.75 & 56.84 & 32.2 & 77.75 & 68.35 & 61.84 \\
GPTQ-b3 & 39.68 & 72.56 & 73.55 & 53.97 & 28.8 & 76.06 & 66.77 & 58.77 \\
% AWQ-b3 & 38.99 & 73.99 & 71.38 & 54.74 & 31 & 76.93 & 66.85 & 59.13 \\
GPTQ-b2 & 20.31 & 29.08 & 50.67 & 27.45 & 12.4 & 54.9 & 51.07 & 35.13 \\
AWQ-b2 & 22.27 & 26.68 & 39.94 & 26.17 & 15.4 & 52.94 & 49.41 & 33.26 \\
OmniQuant-b2 & 20.73 & 38.8 & N/A & 30.11 & N/A & 57.34 & 51.78 & N/A\\
QUIP-b2 & 18.94 & 28.7 & N/A & 27.52 & N/A & 56.53 & 48.78 & N/A \\
QUIP\#-b2 & 28.84 & 55.56 & N/A & 42.94 & N/A & 71.38 & 62.43 & N/A \\
AQLM-b2 & 32.76 & 63.68 & N/A & 49.55 & N/A & 74.76 & 65.67 & N/A \\
AQLM*-b2 & 34.9 & 66.5 & N/A & 50.88 & N/A & 74.92 & 62.43 & N/A \\
SliM-LLM-b2 & 23.38 & 47.81 & 59.97 & 33.76 & 17.8 & 63.82 & 56.91 & 43.35 \\
% Bi-LLM & 21.16 & 39.94 & N/A & 30.74 & N/A & 60.39 & 51.93 &  \\
PB-LLM-b1.7 & 19.11 & 28.2 & 41.25 & 27.03 & 13.2 & 54.46 & 49.09 & 33.19 \\
PTQ1.61-b1.61 & 22.27 & 47.18 & N/A & 35.78 & N/A & 63.22 & 52.25 & N/A \\
% PT2-LLM & 21.08 & 47.01 & 62.91 & 33.82 & 18.8 & 62.95 & 56.75 & 43.33 \\
\textbf{PTQTP-b1.58-D$^\dagger$} & \underline{42.24} & \underline{74.45} & \underline{74.46} & \underline{54.89} & \underline{31.4} & \underline{77.26} & \underline{68.43} & \underline{60.45} \\
\midrule
\multicolumn{9}{c}{\textbf{Qwen3-14B-instruct}} \\
\midrule
FP16 & 55.8 & 83.5 & 86.67 & 61.75 & 35.2 & 80.52 & 74.11 & 68.22 \\
\midrule
AWQ-b4 & \textbf{53.84} & \textbf{81.61} & \textbf{85.6} & \underline{59.13} & \textbf{33.2} & \textbf{79.71} & \textbf{72.38} & \textbf{66.49} \\
GPTQ-b4 & 20.99 & 25.38 & 38.44 & 26.01 & 14.4 & 52.45 & 50.91 & 32.65 \\
% QTIP-b4 &  &  &  &  &  &  &  &  \\
AWQ-b3 & 40.19 & 65.36 & 71.13 & 42.5 & 31.8 & 69.75 & 60.85 & 54.51 \\
% GPTQ-b3 & 22.53 & 25.42 & 38.26 & 25.8 & 13.8 & 53.43 & 50.36 & 32.8 \\
AWQ-b2 & 21.59 & 25.17 & 42.14 & 25.73 & 16.2 & 53.81 & 52.01 & 33.81 \\
GPTQ-b2 & 22.61 & 24.96 & 38.17 & 25.57 & 13.2 & 52.07 & 50.28 & 32.41 \\
SliM-LLM-b2 & 29.35 & 52.54 & 61.2 & 31.52 & 20.4 & 61.83 & 52.04 & 44.13 \\
PB-LLM-b1.7 & 20.73 & 25.93 & 38.04 & 25.76 & 15 & 54.08 & 47.99 & 32.5 \\
PT2-LLM & 23.63 & 53.03 & 62.17 & 33.65 & 20.6 & 62.95 & 59.75 & 45.11 \\
\textbf{PTQTP-b1.58-D$^\dagger$} & \underline{53.5} & \underline{81.44} & \underline{75.5} & \textbf{60.27} & \underline{32.2} & \underline{78.73} & \underline{71.35} & \underline{64.71} \\
\bottomrule
\end{tabular}%
% }
\end{table*}

%% file: tables/newtable3_benchmarkQAT.tex
\begin{table*}[t]
% \scriptsize
\small
\setlength{\tabcolsep}{2pt}
\renewcommand{\arraystretch}{1.3} % Adjusted for better cell padding
\centering
\caption{Complex tasks comparison between PTQTP-quantized models and leading instruction-tuned LLMs (1B-4B parameters)/ SOTA PTQ-quantized models across efficiency metrics and benchmark performance (accuracy \%). D$^\dagger$ : Dual. \textbf{Bold}: best result; \underline{underlined}: second-best. * Results claimed by BitNet-b1.58 paper.}
\label{tab_compare_qat}
% \resizebox{\linewidth}{!}{%

\begin{tabular}{@{}l*{6}{c}@{}}

\toprule
\shortstack{\textbf{Model (Params)}} &
\shortstack{\textbf{Math-500}} &
\shortstack{\textbf{GSM8K}} &
\shortstack{\textbf{HumanEval}} &
\shortstack{\textbf{MBPP}} &
\shortstack{\textbf{MMLU}} &
\shortstack{\textbf{Storage/GB}}\\
% \shortstack{\textbf{AVG}} \\
\midrule
% ------------------------------------------------------------------
% ------------------------------------------------------------------
% MIN/MAX for this table's columns (PLEASE VERIFY THESE VALUES!)
% Math-500:  Min = 0.60 , Max = 82.60
% GSM8K:     Min = 46.47 , Max = 87.79
% HumanEval:  Min = 17.68 , Max = 65.24
% MBPP:     Min = 44.75 , Max = 63.81
% MMLU:      Min =  43.82 , Max = 63.65
% AVG:  Min = 49.33 , Max = 74.25

% ------------------------------------------------------------------
LLaMA3.2 (1B)
& \applyECTGradedColor{14.40}{14.40}{82.60}{14.40}
& \applyECTGradedColor{46.47}{46.47}{87.79}{46.47}

& \applyECTGradedColor{41.46}{17.68}{65.24}{41.46}
& \applyECTGradedColor{52.14}{44.75}{63.81}{52.14}

& \applyECTGradedColor{46.81}{43.82}{63.65}{46.81}
& 2.74 \\

% Qwen2.5 (1.5B)
% & \applyECTGradedColor{56.20}{14.40}{82.60}{56.20}
% & \applyECTGradedColor{64.67}{46.47}{87.79}{64.67}

% & \applyECTGradedColor{53.66}{17.68}{65.24}{53.66}
% & \applyECTGradedColor{57.59}{44.75}{63.81}{57.59}

% & \applyECTGradedColor{61.91}{43.82}{63.65}{61.91}
% & 3.09 \\
% % & \applyECTGradedColor{72.39}{49.33}{74.25}{72.39}\\

Qwen3 (1.7B)
& \applyECTGradedColor{72.20}{14.40}{82.60}{{72.20}}
& \applyECTGradedColor{74.60}{46.47}{87.79}{{74.60}}

& \applyECTGradedColor{65.24}{17.68}{65.24}{65.24}
& \applyECTGradedColor{63.42}{44.75}{63.81}{63.42}

& \applyECTGradedColor{60.63}{43.82}{63.65}{60.63}
& 4.04 \\
% & \applyECTGradedColor{69.54}{49.33}{74.25}{69.54}\\

SmolLM2 (1.7B)
& \applyECTGradedColor{21.00}{0.6}{82.60}{21.00}
& \applyECTGradedColor{50.19}{46.47}{87.79}{50.19}

& \applyECTGradedColor{35.98}{17.68}{65.24}{35.98}
& \applyECTGradedColor{49.81}{44.75}{63.81}{49.81}

& \applyECTGradedColor{49.73}{43.82}{63.65}{49.73}
& 3.42 \\
% & \applyECTGradedColor{60.07}{49.33}{74.25}{60.07}\\

MiniCPM (2B)
& \applyECTGradedColor{0.60}{0.6}{82.60}{0.60}
& \applyECTGradedColor{56.48}{46.47}{87.79}{56.48}

& \applyECTGradedColor{39.02}{17.68}{65.24}{39.02}
& \applyECTGradedColor{55.64}{44.75}{63.81}{55.64}

& \applyECTGradedColor{52.08}{43.82}{63.65}{52.08}
&5.45 \\
% & \applyECTGradedColor{50.45}{49.33}{74.25}{50.45}\\

BitNet-b1.58 (2B)$^{*}$
& \applyECTGradedColor{43.40}{14.40}{82.60}{43.40}
& \applyECTGradedColor{58.38}{46.47}{87.79}{58.38}

% & \applyECTGradedColor{--}{17.68}{65.24}{--}
% & \applyECTGradedColor{--}{44.75}{63.81}{--}

& N/A
& N/A

& \applyECTGradedColor{53.17}{43.82}{63.65}{53.17}
& 1.18 \\
% & \applyECTGradedColor{66.27}{49.33}{74.25}{66.27}\\

% Qwen2.5 (3B)
% & \applyECTGradedColor{42.80}{14.40}{82.60}{42.80}
% & \applyECTGradedColor{62.47}{46.47}{87.79}{62.47}
% & \applyECTGradedColor{75.93}{49.91}{82.71}{75.93}
% & \applyECTGradedColor{85.54}{22.71}{89.59}{85.54}
% & \applyECTGradedColor{78.38}{26.46}{84.13}{78.38}
% & \applyECTGradedColor{60.87}{37.92}{79.76}{60.87}
% & \applyECTGradedColor{66.32}{26.49}{77.09}{66.32}
% & \applyECTGradedColor{56.80}{43.82}{63.65}{56.80}
% & \applyECTGradedColor{56.27}{50.04}{71.90}{\underline{56.27}} \\

LLaMA3.2 (3B)
& \applyECTGradedColor{45.20}{14.40}{82.60}{45.20}
& \applyECTGradedColor{77.56}{46.47}{87.79}{{77.56}}

& \applyECTGradedColor{62.20}{17.68}{65.24}{62.20}
& \applyECTGradedColor{63.81}{44.75}{63.81}{63.81}

& \applyECTGradedColor{62.24}{43.82}{63.65}{{62.24}}
& 5.98 \\
% & \applyECTGradedColor{73.13}{49.33}{74.25}{\underline{73.13}}\\

Qwen3 (4B)
& \applyECTGradedColor{45.20}{14.40}{82.60}{\textbf{83.00}}
& \applyECTGradedColor{77.56}{46.47}{87.79}{\underline{86.96}}

& \applyECTGradedColor{62.20}{17.68}{65.24}{\textbf{77.44}}
& \applyECTGradedColor{63.81}{44.75}{63.81}{\textbf{79.38}}
& \applyECTGradedColor{62.24}{43.82}{63.65}{\textbf{70.67}}
& 7.49 \\
\midrule
% ----------------- 下面部分 -----------------

Qwen3-PTQTP-b1.58-D$^\dagger$ (1.7B)
& \applyECTGradedColor{43.80}{14.40}{82.60}{43.80}
& \applyECTGradedColor{54.21}{46.47}{87.79}{54.21}

& \applyECTGradedColor{45.53}{17.68}{65.24}{45.53}
& \applyECTGradedColor{48.78}{44.75}{63.81}{48.78}

& \applyECTGradedColor{43.82}{43.82}{63.65}{43.82}
& 1.90\\
% & \applyECTGradedColor{49.33}{49.33}{74.25}{49.33}\\

Qwen2.5-PTQTP-b1.58-D$^\dagger$ (3B)
& \applyECTGradedColor{42.80}{14.40}{82.60}{42.80}
& \applyECTGradedColor{62.47}{46.47}{87.79}{62.47}

& \applyECTGradedColor{25.61}{17.68}{65.24}{25.61}
& \applyECTGradedColor{44.75}{44.75}{63.81}{44.75}

& \applyECTGradedColor{56.80}{43.82}{63.65}{56.80}
& 2.61 \\
% & \applyECTGradedColor{70.55}{49.33}{74.25}{70.55}\\

LLaMA3.2-PTQTP-b1.58-D$^\dagger$ (3B)
& \applyECTGradedColor{34.20}{14.40}{82.60}{34.20}
& \applyECTGradedColor{65.81}{46.47}{87.79}{65.81}

& \applyECTGradedColor{17.68}{17.68}{65.24}{17.68}
& \applyECTGradedColor{52.92}{44.75}{63.81}{52.92}

& \applyECTGradedColor{55.08}{43.82}{63.65}{55.08}
& 2.94\\
% & \applyECTGradedColor{67.34}{49.33}{74.25}{67.34}\\

Qwen3-PTQTP-b1.58-D$^\dagger$ (4B)
& \applyECTGradedColor{82.60}{14.40}{82.60}{\underline{82.60}}
& \applyECTGradedColor{87.79}{46.47}{87.79}{\textbf{87.79}}

& \applyECTGradedColor{71.34}{17.68}{71.34}{\underline{71.34}}
& \applyECTGradedColor{69.26}{44.75}{69.26}{\underline{69.26}}

& \applyECTGradedColor{63.65}{43.82}{63.65}{\underline{63.65}}
& 3.35 \\
% & \applyECTGradedColor{74.25}{49.33}{74.25}{\textbf{74.93}}\\

\midrule

% LLaMA3 (8B)
% & \applyECTGradedColor{27.80}{14.40}{82.60}{27.80}
% & \applyECTGradedColor{79.23}{46.47}{87.79}{79.23}

% & \applyECTGradedColor{59.15}{17.68}{65.24}{59.15}
% & \applyECTGradedColor{--}{44.75}{63.81}{--}

% & \applyECTGradedColor{--}{43.82}{63.65}{--}
% & \applyECTGradedColor{--}{49.33}{74.25}{--}\\

% LLaMA3-AWQ-b2 (8B)
% & \applyECTGradedColor{0.00}{14.40}{82.60}{0.00}
% & \applyECTGradedColor{0.15}{46.47}{87.79}{0.15}

% & \applyECTGradedColor{0.00}{17.68}{65.24}{0.00}
% & \applyECTGradedColor{--}{44.75}{63.81}{--}

% & \applyECTGradedColor{--}{43.82}{63.65}{--}
% & \applyECTGradedColor{--}{49.33}{74.25}{--}\\

% LLaMA3-AQLM-b2.07 (8B)
% & \applyECTGradedColor{12.40}{14.40}{82.60}{12.40}
% & \applyECTGradedColor{60.42}{46.47}{87.79}{60.42}

% & \applyECTGradedColor{28.66}{17.68}{65.24}{28.66}
% & \applyECTGradedColor{--}{44.75}{63.81}{--}

% & \applyECTGradedColor{--}{43.82}{63.65}{--}
% & \applyECTGradedColor{--}{49.33}{74.25}{--}\\

% LLaMA3-PTQTP-b1.58-D$^\dagger$ (8B)
% & \applyECTGradedColor{15.40}{14.40}{82.60}{15.40}
% & \applyECTGradedColor{67.02}{46.47}{87.79}{67.02}

% & \applyECTGradedColor{42.07}{17.68}{65.24}{42.07}
% & \applyECTGradedColor{--}{44.75}{63.81}{--}

% & \applyECTGradedColor{--}{43.82}{63.65}{--}
% & \applyECTGradedColor{--}{49.33}{74.25}{--}\\

LLaMA3 (8B)
& \applyECTGradedColor{27.80}{14.40}{82.60}{27.80}
& \applyECTGradedColor{79.23}{46.47}{87.79}{79.23}

& \applyECTGradedColor{59.15}{17.68}{65.24}{59.15}
& \applyECTGradedColor{68.48}{44.75}{63.81}{68.48}

& \applyECTGradedColor{68.28}{43.82}{63.65}{68.28}
& 14.96 \\
% & \applyECTGradedColor{60.59}{49.33}{74.25}{60.59}\\

LLaMA3-AWQ-b2 (8B)
& \applyECTGradedColor{0.00}{14.40}{82.60}{0.00}
& \applyECTGradedColor{0.15}{46.47}{87.79}{0.15}

& \applyECTGradedColor{0.00}{17.68}{65.24}{0.00}
& \applyECTGradedColor{0.00}{44.75}{63.81}{0.00}

& \applyECTGradedColor{24.82}{43.82}{63.65}{24.82}
& 2.86 \\
% & \applyECTGradedColor{4.99}{49.33}{74.25}{4.99}\\

LLaMA3-AQLM-b2.07 (8B)
& \applyECTGradedColor{12.40}{14.40}{82.60}{12.40}
& \applyECTGradedColor{60.42}{46.47}{87.79}{60.42}

& \applyECTGradedColor{28.66}{17.68}{65.24}{28.66}
& \applyECTGradedColor{29.27}{44.75}{63.81}{29.27}

& \applyECTGradedColor{47.86}{43.82}{63.65}{47.86}
& 4.08 \\
% & \applyECTGradedColor{35.72}{49.33}{74.25}{35.72}

LLaMA3-PTQTP-b1.58-D$^\dagger$ (8B)
& \applyECTGradedColor{15.40}{14.40}{82.60}{15.40}
& \applyECTGradedColor{67.02}{46.47}{87.79}{67.02}

& \applyECTGradedColor{42.07}{17.68}{65.24}{42.07}
& \applyECTGradedColor{45.91}{44.75}{63.81}{45.91}

& \applyECTGradedColor{54.54}{43.82}{63.65}{54.54}
& 5.61 \\
% & \applyECTGradedColor{44.99}{49.33}{74.25}{44.99}\\

% ------------------------------------------------------------------

\bottomrule

\end{tabular}%
% }

\end{table*}

%% file: tables/table_add_revised.tex
\begin{table}[tbp]
\centering
\small
\setlength{\tabcolsep}{2pt}
\caption{{Perplexity (PPL) of LLaMA-3.1 8B and LLaMA-3.2 3B on WikiText-2, PTB, and C4 under different regularisation strengths. Lower is better.}}
\label{tab:ppl}
% \resizebox{\linewidth}{!}{%
\begin{tabular}{c|ccc|ccc}
\toprule
\multirow{2}{*}{Condition} & \multicolumn{3}{c|}{LLaMA-3.1 8B} & \multicolumn{3}{c}{LLaMA-3.2 3B} \\
\cmidrule(lr){2-4}\cmidrule(lr){5-7}
 & WikiText2 & PTB & C4 & WikiText2 & PTB & C4 \\
\midrule
1            & 12.45 & 17.52 & 17.76 & 12.52 & 21.20 & 18.09 \\
5            & 10.86 & 16.11 & 16.17 & 11.52 & 19.98 & 17.13 \\
$10^1$       & 9.58  & 14.92 & 14.65 & 10.92 & 18.64 & 16.18 \\
$10^2$       & 8.97  & 14.20 & 13.81 & 10.52 & 17.76 & 15.56 \\
$10^4$       & 8.62  & 13.84 & 13.40 & 10.26 & 17.32 & 15.26 \\
% $10^8$       & 8.53  & 13.72 & 13.22 & 10.24 & 17.41 & 15.28 \\
% $10^2$       & 8.53  & 13.72 & 13.22 & 10.24 & 17.41 & 15.28 \\
$10^6$       & 8.53  & 13.72 & 13.22 & 10.24 & 17.41 & 15.28 \\
$10^{12}$    & 8.53  & 13.72 & 13.22 & 10.24 & 17.41 & 15.28 \\
% $10^{18}$    & 8.53  & 13.72 & 13.22 & 10.24 & 17.41 & 15.28 \\
\bottomrule
\end{tabular}
% }
\end{table}

%% file: algs/arg.tex
\begin{algorithm}[t]
\caption{Progressive and Adaptive Regularization}
\label{alg_trit_decomp}
\begin{algorithmic}[1]
\Require Weight matrix $W \in \mathbb{R}^{n \times d}$, max iterations $T_{\text{max}}$, tolerance $\epsilon$
\Ensure Optimized parameters $\alpha, T^{(1)}, T^{(2)}$

\State \textbf{Initialize:}
\State $T^{(k)}_{(0)} \gets \text{sign}(W)$ with $0 \to 1$ replacement for $k=1,2$ %\Comment{Initial trit-planes}
\State Let $\alpha_{(0)} \gets [1, 1] \otimes \mathbf{1}_n$ %\Comment{Uniform initial scaling coefficients}
\State Let $\lambda_{(0)} \gets 10^{-8} \cdot \mathbf{1}_n$ %\Comment{Initial regularization parameters}

\For{$t = 1$ \textbf{to} $T_{\text{max}}$}
    % \State \textbf{Update Continuous Scaling Coefficients:}
    \For{row $i = 1$ \textbf{to} $n$}
        \State $S_i \gets [T^{(1)T}_{i,(t-1)} \quad T^{(2)T}_{i,(t-1)}]$ %\Comment{Local basis matrix}
        \State $A_i \gets (S_i)^T S_i + \lambda_i I_2$
        \State $b_i \gets (S_i)^T W_i^T$
        \State Solve $\alpha_{(i, t)} \gets (A_i)^{-1} b_i$ %\Comment{Ridge regression solution}
        \State $\lambda_{i, new} \gets \lambda_{i} \sqrt{{\kappa_{i,approx}/{10^{6}}}}$, 
        \State When $\kappa_{i, approx} \geq 10^{6}$
    \EndFor
    
    % \State \textbf{Optimize Discrete Trit-Planes:}
    \State $\mathcal{C} \gets \{-1, 0, 1\}^2$ %\Comment{All possible ternary value pairs}
    \For{element $(i,j) \in W$}
        \State Evaluate error for all $\mathbf{c}_m \in \mathcal{C}$: 
        \State \quad $e_m \gets \left(W_{ij} - \sum_{k=1}^2 \alpha_{i,(t)}^{(k)} c^{(k)}_{m}\right)^2$
        \State $m^* \gets \arg\min_m e_m$ %\Comment{Select best ternary pair}
        \State $T^{(1)}_{ij,(t)}, T^{(2)}_{ij,(t)} \gets \mathbf{c}^*_{m}$ %\Comment{Update trit-plane elements}
    \EndFor
    
    % \State \textbf{Convergence Check:}
    \If{$\|\alpha_{(t)} - \alpha_{(t-1)}\|_F < \epsilon$}
        \State \textbf{break} %\Comment{Early termination if converged}
    \EndIf
\EndFor

\State \textbf{Return} $\alpha_{(t)}, \{T^{(1)}_{(t)}, T^{(2)}_{(t)}\}$
\end{algorithmic}
\end{algorithm}

%% file: tables/Downstream_qwen3.tex
\begin{table}[t]
% \scriptsize
\small
\setlength{\tabcolsep}{2pt}
\renewcommand{\arraystretch}{0.85}
\centering
\caption{Accuracy of 7 language reasoning tasks on Qwen3 family with PTQTP. We compare the results among Qwen‑3 Models FP16 and PTQTP across model size from 0.6B to 32B. HellaS: HellaSwag. \textbf{Bold}: Keep original performance more than 95\% after using PTQTP.}
\label{tab_qwen3-FP16-PTQTP}
% \resizebox{\linewidth}{!}{%
\begin{tabular}{llcccccc}
\toprule
\multirow{2}{*}{\textbf{Metrics}} & \multirow{2}{*}{\textbf{Method}} & \multicolumn{6}{c}{\textbf{Size}} \\
\cmidrule(lr){3-8}
& & 0.6B & 1.7B & 4B & 8B & 14B & 32B \\
\midrule

\multirow{3}{*}{ARC‑C}
& FP16 & 52.88 & 65.08 & 79.66 & 82.37 & 82.37 & 88.81 \\
& PTQTP & 44.41 & 51.53 & 76.95 & 68.81 & 80.68 & 77.97 \\
% & \textcolor{green!50!black}&{(\%)} & \textcolor{green!50!black}{84.0} & \textcolor{green!50!black}{79.2} & \textcolor{green!50!black}{96.6} & \textcolor{green!50!black}{83.5} & \textcolor{green!50!black}{97.9} & \textcolor{green!50!black}{87.8} \\
&\cellcolor{lightgray!40} {(\%)} & \cellcolor{lightgray!40}{84.0} & \cellcolor{lightgray!40}{79.2} & \cellcolor{lightgray!40}{\textbf{96.6}} & \cellcolor{lightgray!40}{83.5} & \cellcolor{lightgray!40}{\textbf{97.9}} & \cellcolor{lightgray!40}{87.8} \\
\midrule

\multirow{3}{*}{ARC‑E}
& FP16 & 73.72 & 80.25 & 87.13 & 89.07 & 90.83 & 92.24 \\
& PTQTP & 56.44 & 70.02 & 86.60 & 81.13 & 88.71 & 86.24 \\
% & \textcolor{green!50!black}&{(\%)} & \textcolor{green!50!black}{76.6} & \textcolor{green!50!black}{87.3} & \textcolor{green!50!black}{99.4} & \textcolor{green!50!black}{91.1} & \textcolor{green!50!black}{97.7} & \textcolor{green!50!black}{93.5} \\
&\cellcolor{lightgray!40} {(\%)} & \cellcolor{lightgray!40}{76.6} &  \cellcolor{lightgray!40}{87.3} &\cellcolor{lightgray!40}  {\textbf{99.4}} &\cellcolor{lightgray!40}  {91.1} &\cellcolor{lightgray!40} {\textbf{97.7}} &  \cellcolor{lightgray!40}{93.5} \\
\midrule

\multirow{3}{*}{BoolQ}
& FP16 & 66.42 & 78.87 & 86.36 & 86.39 & 89.45 & 87.98 \\
& PTQTP & 62.57 & 27.55 & 84.13 & 85.44 & 89.02 & 86.27 \\
% & \textcolor{green!50!black}&{(\%)} & \textcolor{green!50!black}{94.2} & \textcolor{green!50!black}{34.9} & \textcolor{green!50!black}{97.4} & \textcolor{green!50!black}{98.9} & \textcolor{green!50!black}{99.5} & \textcolor{green!50!black}{98.1} \\
&\cellcolor{lightgray!40} {(\%)} & \cellcolor{lightgray!40} {94.2} & \cellcolor{lightgray!40} {34.9} & \cellcolor{lightgray!40} {\textbf{97.4}} & \cellcolor{lightgray!40} {\textbf{98.9}} & \cellcolor{lightgray!40} {\textbf{99.5}} & \cellcolor{lightgray!40} {\textbf{98.1}} \\
\midrule

\multirow{3}{*}{HellaS}
& FP16 & 43.60 & 58.03 & 78.19 & 84.67 & 88.09 & 90.80 \\
& PTQTP & 32.85 & 37.92 & 71.02 & 79.75 & 85.13 & 89.56 \\
% & \textcolor{green!50!black}&{(\%)} & \textcolor{green!50!black}{75.3} & \textcolor{green!50!black}{65.3} & \textcolor{green!50!black}{90.8} & \textcolor{green!50!black}{94.2} & \textcolor{green!50!black}{96.6} & \textcolor{green!50!black}{98.6} \\
&\cellcolor{lightgray!40}{(\%)} &  \cellcolor{lightgray!40} {75.3} & \cellcolor{lightgray!40}  {65.3} & \cellcolor{lightgray!40}  {90.8} & \cellcolor{lightgray!40}   {94.2} & \cellcolor{lightgray!40}  {\textbf{96.6}} &  \cellcolor{lightgray!40} {\textbf{98.6}} \\
\midrule

\multirow{3}{*}{PIQA}
& FP16 & 58.98 & 67.57 & 77.26 & 80.20 & 81.07 & 83.51 \\
& PTQTP & 57.67 & 57.89 & 73.18 & 73.78 & 79.33 & 80.41 \\
% & \textcolor{green!50!black}&{(\%)} & \textcolor{green!50!black}{97.8} & \textcolor{green!50!black}{85.7} & \textcolor{green!50!black}{94.7} & \textcolor{green!50!black}{92.0} & \textcolor{green!50!black}{97.9} & \textcolor{green!50!black}{96.3} \\
 &\cellcolor{lightgray!40} {(\%)} &  \cellcolor{lightgray!40}  {\textbf{97.8}} &   \cellcolor{lightgray!40} {85.7} &  \cellcolor{lightgray!40}  {\textbf{94.7}} & \cellcolor{lightgray!40}   {92.0} &  \cellcolor{lightgray!40}  {\textbf{97.9}} &  \cellcolor{lightgray!40}  {\textbf{96.3}} \\
\midrule

\multirow{3}{*}{WinoG}
& FP16 & 50.51 & 50.99 & 58.96 & 54.14 & 68.90 & 77.82 \\
& PTQTP & 50.43 & 51.07 & 53.67 & 59.67 & 67.72 & 73.64 \\
% & \textcolor{green!50!black}&{(\%)} & \textcolor{green!50!black}{99.8} & \textcolor{green!50!black}{100.2} & \textcolor{green!50!black}{91.0} & \textcolor{green!50!black}{110.2} & \textcolor{green!50!black}{98.3} & \textcolor{green!50!black}{94.6} \\
  &\cellcolor{lightgray!40} {(\%)} &   \cellcolor{lightgray!40}  {\textbf{99.8}} &    \cellcolor{lightgray!40} {\textbf{100.2}} &  \cellcolor{lightgray!40}  {91.0} &  \cellcolor{lightgray!40}   {\textbf{110.2}} &  \cellcolor{lightgray!40}   {\textbf{98.3}} &   \cellcolor{lightgray!40}  {94.6} \\
\midrule

\multirow{3}{*}{MMLU}
& FP16 & 43.76 & 60.67 & 70.68 & 76.55 & 79.38 & 82.81 \\
& PTQTP & 33.64 & 43.82 & 63.65 & 68.23 & 76.20 & 80.56 \\
% & \textcolor{green!50!black}&{(\%)} & \textcolor{green!50!black}{76.9} & \textcolor{green!50!black}{72.2} & \textcolor{green!50!black}{90.1} & \textcolor{green!50!black}{89.1} & \textcolor{green!50!black}{96.0} & \textcolor{green!50!black}{97.3} \\
   & \cellcolor{lightgray!40}{(\%)} &   \cellcolor{lightgray!40}   {76.9} &    \cellcolor{lightgray!40}  {72.2} &  \cellcolor{lightgray!40}    {90.1} &  \cellcolor{lightgray!40}    {89.1} &    \cellcolor{lightgray!40}  {\textbf{96.0}} &   \cellcolor{lightgray!40}   {\textbf{97.3}} \\
\bottomrule
\end{tabular}%
\vspace{-0.5em}
% }
\end{table}

%% file: tables/benchmark_all.tex
\begin{table*}[t]
% \scriptsize
\small
\setlength{\tabcolsep}{2pt}
\renewcommand{\arraystretch}{1.3} % Adjusted for better cell padding
\centering
\caption{Comparison between PTQTP-quantized models and leading instruction-tuned LLMs (1B-4B parameters) across efficiency metrics and benchmark performance (accuracy \%). D$^\dagger$ : Dual. \textbf{Bold}: best result; \underline{underlined}: second-best. * Results claimed by BitNet-b1.58 paper. }
\label{tab_compare_qat_more}
% \resizebox{\linewidth}{!}{%

\begin{tabular}{@{}l*{6}{c}@{}}

\toprule
\shortstack{\textbf{Model (Params)}} &
% \shortstack{\textbf{Math-500}} &
% \shortstack{\textbf{GSM8K}} &
\shortstack{\textbf{ARC-C}} &
\shortstack{\textbf{ARC-E}} &
\shortstack{\textbf{BoolQ}} &
\shortstack{\textbf{HellaSwag}} &
\shortstack{\textbf{PIQA}} &
% \shortstack{\textbf{MMLU}} &
\shortstack{\textbf{WinoG}} \\
\midrule
% ------------------------------------------------------------------
% ------------------------------------------------------------------
% MIN/MAX for this table's columns (PLEASE VERIFY THESE VALUES!)
% Math-500:  Min = 14.40 , Max = 82.60
% GSM8K:     Min = 46.47 , Max = 87.79
% ARC-C:     Min = 49.91 , Max = 82.71
% ARC-E:     Min = 22.71 , Max = 89.59
% BoolQ:     Min = 26.46 , Max = 84.13
% HellaSwag: Min = 37.92 , Max = 79.76
% PIQA:      Min = 26.49 , Max = 77.09
% MMLU:      Min =  0.60 , Max = 63.65
% WinoG:     Min = 50.04 , Max = 71.90
% ------------------------------------------------------------------
LLaMA3.2 (1B)
% & \applyECTGradedColor{14.40}{14.40}{82.60}{14.40}
% & \applyECTGradedColor{46.47}{46.47}{87.79}{46.47}
& \applyECTGradedColor{55.93}{49.91}{82.71}{55.93}
& \applyECTGradedColor{69.84}{22.71}{89.59}{69.84}
& \applyECTGradedColor{62.20}{26.46}{84.13}{62.20}
& \applyECTGradedColor{40.12}{37.92}{79.76}{40.12}
& \applyECTGradedColor{58.16}{26.49}{77.09}{58.16}
% & \applyECTGradedColor{46.81}{43.82}{63.65}{46.81}
& \applyECTGradedColor{50.04}{50.04}{71.90}{50.04} \\

% Qwen2.5 (1.5B)
% % & \applyECTGradedColor{56.20}{14.40}{82.60}{56.20}
% % & \applyECTGradedColor{64.67}{46.47}{87.79}{64.67}
% & \applyECTGradedColor{71.86}{49.91}{82.71}{71.86}
% & \applyECTGradedColor{89.59}{22.71}{89.59}{\textbf{89.59}}
% & \applyECTGradedColor{79.48}{26.46}{84.13}{79.48}
% & \applyECTGradedColor{61.88}{37.92}{79.76}{61.88}
% & \applyECTGradedColor{75.73}{26.49}{77.09}{\underline{75.73}}
% % & \applyECTGradedColor{61.91}{43.82}{63.65}{61.91}
% & \applyECTGradedColor{55.80}{50.04}{71.90}{55.80} \\

Qwen3 (1.7B)
% & \applyECTGradedColor{72.20}{14.40}{82.60}{\underline{72.20}}
% & \applyECTGradedColor{74.60}{46.47}{87.79}{{74.60}}
& \applyECTGradedColor{82.71}{49.91}{82.71}{\textbf{82.71}}
& \applyECTGradedColor{80.07}{22.71}{89.59}{80.07}
& \applyECTGradedColor{78.59}{26.46}{84.13}{78.59}
& \applyECTGradedColor{57.76}{37.92}{79.76}{57.76}
& \applyECTGradedColor{67.36}{26.49}{77.09}{67.36}
% & \applyECTGradedColor{60.63}{43.82}{63.65}{60.63}
& \applyECTGradedColor{50.75}{50.04}{71.90}{50.75} \\

SmolLM2 (1.7B)
% & \applyECTGradedColor{21.00}{0.6}{82.60}{21.00}
% & \applyECTGradedColor{50.19}{46.47}{87.79}{50.19}
& \applyECTGradedColor{49.15}{49.91}{82.71}{49.15}
& \applyECTGradedColor{76.01}{22.71}{89.59}{76.01}
& \applyECTGradedColor{67.25}{26.46}{84.13}{67.25}
& \applyECTGradedColor{50.41}{37.92}{71.02}{50.41}
& \applyECTGradedColor{67.57}{53.75}{77.09}{67.57}
% & \applyECTGradedColor{49.73}{43.82}{63.65}{49.73}
& \applyECTGradedColor{50.04}{50.04}{71.90}{50.04} \\

MiniCPM (2B)
% & \applyECTGradedColor{0.60}{0.6}{82.60}{0.60}
% & \applyECTGradedColor{56.48}{46.47}{87.79}{56.48}
& \applyECTGradedColor{62.03}{49.91}{82.71}{62.03}
& \applyECTGradedColor{26.46}{22.71}{89.59}{26.46}
& \applyECTGradedColor{79.76}{26.46}{84.13}{79.76}
& \applyECTGradedColor{26.49}{37.92}{71.02}{26.49}
& \applyECTGradedColor{53.75}{53.75}{77.09}{53.75}
% & \applyECTGradedColor{52.08}{43.82}{63.65}{52.08}
& \applyECTGradedColor{54.22}{50.04}{71.90}{54.22} \\

BitNet-b1.58 (2B)$^{*}$
% & \applyECTGradedColor{43.40}{14.40}{82.60}{43.40}
% & \applyECTGradedColor{58.38}{46.47}{87.79}{58.38}
& \applyECTGradedColor{49.91}{49.91}{82.71}{49.91}
& \applyECTGradedColor{74.79}{22.71}{89.59}{74.79}
& \applyECTGradedColor{80.18}{26.46}{84.13}{80.18}
& \applyECTGradedColor{68.44}{37.92}{79.76}{\underline{68.44}}
& \applyECTGradedColor{77.09}{26.49}{77.09}{\textbf{77.09}}
% & \applyECTGradedColor{53.17}{43.82}{63.65}{53.17}
& \applyECTGradedColor{71.90}{50.04}{71.90}{\textbf{71.90}} \\

% Qwen2.5 (3B)
% & \applyECTGradedColor{42.80}{14.40}{82.60}{42.80}
% & \applyECTGradedColor{62.47}{46.47}{87.79}{62.47}
% & \applyECTGradedColor{75.93}{49.91}{82.71}{75.93}
% & \applyECTGradedColor{85.54}{22.71}{89.59}{85.54}
% & \applyECTGradedColor{78.38}{26.46}{84.13}{78.38}
% & \applyECTGradedColor{60.87}{37.92}{79.76}{60.87}
% & \applyECTGradedColor{66.32}{26.49}{77.09}{66.32}
% & \applyECTGradedColor{56.80}{43.82}{63.65}{56.80}
% & \applyECTGradedColor{56.27}{50.04}{71.90}{\underline{56.27}} \\

LLaMA3.2 (3B)
% & \applyECTGradedColor{45.20}{14.40}{82.60}{45.20}
% & \applyECTGradedColor{77.56}{46.47}{87.79}{\underline{77.56}}
& \applyECTGradedColor{80.68}{49.91}{82.71}{\underline{80.68}}
& \applyECTGradedColor{88.36}{22.71}{89.59}{\textbf{88.36}}
& \applyECTGradedColor{80.89}{26.46}{84.13}{\underline{80.89}}
& \applyECTGradedColor{63.75}{37.92}{79.76}{63.75}
& \applyECTGradedColor{71.38}{26.49}{77.09}{71.38}
% & \applyECTGradedColor{62.24}{43.82}{63.65}{\underline{62.24}}
& \applyECTGradedColor{53.75}{50.04}{71.90}{53.75} \\

\midrule
% ----------------- 下面部分 -----------------

Qwen3-PTQTP-b1.58-D$^\dagger$ (1.7B)
% & \applyECTGradedColor{43.80}{14.40}{82.60}{43.80}
% & \applyECTGradedColor{54.21}{46.47}{87.79}{54.21}
& \applyECTGradedColor{51.53}{49.91}{82.71}{51.53}
& \applyECTGradedColor{70.02}{22.71}{89.59}{70.02}
& \applyECTGradedColor{27.55}{26.46}{84.13}{27.55}
& \applyECTGradedColor{37.92}{37.92}{79.76}{37.92}
& \applyECTGradedColor{57.89}{26.49}{77.09}{57.89}
% & \applyECTGradedColor{43.82}{43.82}{63.65}{43.82}
& \applyECTGradedColor{51.07}{50.04}{71.90}{51.07} \\

Qwen2.5-PTQTP-b1.58-D$^\dagger$ (3B)
% & \applyECTGradedColor{42.80}{14.40}{82.60}{42.80}
% & \applyECTGradedColor{62.47}{46.47}{87.79}{62.47}
& \applyECTGradedColor{75.93}{49.91}{82.71}{75.93}
& \applyECTGradedColor{85.54}{22.71}{89.59}{85.54}
& \applyECTGradedColor{78.38}{26.46}{84.13}{78.38}
& \applyECTGradedColor{60.87}{37.92}{79.76}{60.87}
& \applyECTGradedColor{66.32}{26.49}{77.09}{66.32}
% & \applyECTGradedColor{56.80}{43.82}{63.65}{56.80}
& \applyECTGradedColor{56.27}{50.04}{71.90}{\underline{56.27}} \\

LLaMA3.2-PTQTP-b1.58-D$^\dagger$ (3B)
% & \applyECTGradedColor{34.20}{14.40}{82.60}{34.20}
% & \applyECTGradedColor{65.81}{46.47}{87.79}{65.81}
& \applyECTGradedColor{74.92}{49.91}{82.71}{74.92}
& \applyECTGradedColor{82.89}{22.71}{89.59}{82.89}
& \applyECTGradedColor{74.34}{26.46}{84.13}{74.34}
& \applyECTGradedColor{55.61}{37.92}{79.76}{55.61}
& \applyECTGradedColor{64.42}{26.49}{77.09}{64.42}
% & \applyECTGradedColor{55.08}{43.82}{63.65}{55.08}
& \applyECTGradedColor{51.85}{50.04}{71.90}{51.85} \\

Qwen3-PTQTP-b1.58-D$^\dagger$ (4B)
% & \applyECTGradedColor{82.60}{14.40}{82.60}{\textbf{82.60}}
% & \applyECTGradedColor{87.79}{46.47}{87.79}{\textbf{87.79}}
& \applyECTGradedColor{76.95}{49.91}{82.71}{76.95}
& \applyECTGradedColor{86.60}{22.71}{89.59}{\underline{86.60}}
& \applyECTGradedColor{84.13}{26.46}{84.13}{\textbf{84.13}}
& \applyECTGradedColor{71.02}{37.92}{71.02}{\textbf{71.02}}
& \applyECTGradedColor{73.18}{26.49}{77.09}{\underline{73.18}}
% & \applyECTGradedColor{63.65}{43.82}{63.65}{\textbf{63.65}}
& \applyECTGradedColor{53.67}{50.04}{71.90}{53.67} \\
% ------------------------------------------------------------------

\bottomrule

\end{tabular}%
% }

\end{table*}

%% file: tables/llamappl.tex
\begin{table*}[tbp]
\centering
\scriptsize
\small
\setlength{\tabcolsep}{2pt}       
\renewcommand{\arraystretch}{0.85} 
\caption{Comparison between PTQTP and binary PTQ methods on LLaMA family across multiple datasets: WikiText2, PTB \citep{ptb}, C4 (Group Size is 128). N/A: No corresponding result; D$^\dagger$: Dual. \textbf{Bold}: best result. }
\label{tab_llama_ppl}
% \resizebox{\linewidth}{!}{%

\begin{tabular}{@{} ll ccc ccc ccc @{}}  % 修正后的列定义
\toprule
\multirow{2}{*}{{Model}} & 
\multirow{2}{*}{{Method/\#Bits}} & 
\multicolumn{3}{c}{{WikiText2}} & 
\multicolumn{3}{c}{{PTB}} & 
\multicolumn{3}{c}{{C4}} \\
\cmidrule(lr){3-5}\cmidrule(lr){6-8}\cmidrule(lr){9-11} 
 &  &7B/8B & 13B & 65B/70B & 7B/8B & 13B & 65B/70B & 7B/8B & 13B & 65B/70B \\
\midrule
\multirow{8}{*}{LLaMA}
 & FP16    & {5.68}  & {5.09}  & {3.53}  & {41.15} & {28.09} & {25.05} & {7.34}  & {6.79}  & {5.81}  \\
 \midrule
 % & AWQ/4.00      & {5.81}  & {5.19}  & {3.62}  & {32.79} & {21.89} & 25.75 & {7.59}  & {6.88}  & {5.90}  \\
 % & AWQ/3.00      & \textbf{6.35}  & \textbf{5.52}  & \textbf{3.95}  & 43.74 & \textbf{25.09} & 50.79 & 9.07  & 7.79  & 6.33  \\
 % & GPTQ/3.00     & 6.55  & 5.67  & 4.17  & 84.88 & 26.40 & {19.55} & 50.31 & 32.31 & \textbf{6.03}  \\
 & GPTQ/2.00     & 129.19& 20.46 & 8.66  & 1421.47&224.45& 47.70 & 79.06 &18.97 &10.23 \\
 & PB-LLM/1.70   & 82.76 & 44.93 &12.81 &603.57 &237.22&119.19 &76.63 &40.64 &15.30 \\
 & BiLLM/1.09    &49.79 &14.58 &8.37  &373.81 &84.87 &44.68  &46.96 &16.83 &11.09 \\
 & ARB-LLM\textsubscript{RC}/1.09    & 14.03 & 10.18 & 6.56 & 195.94& 54.38& 32.20  & 17.38 & 12.48& 8.91 \\
\rowcolor{lightgray!40} & {PTQTP-b1.58-D$^\dagger$}    &  \textbf{6.40} & \textbf{5.66} & \textbf{4.04} & \textbf{50.41} & \textbf{30.25} & \textbf{24.67} & \textbf{8.32} & \textbf{7.41} & \textbf{6.28} \\
\midrule

\multirow{8}{*}{LLaMA-2}
 & FP16    & {5.47}  &{4.88}  &{3.32}  &{37.9}  &{50.93} &{24.25} &{7.26}  &{6.73}  &{5.71}  \\
 \midrule
 % & AWQ/4.00      &{5.61}  &{4.97}  &{3.41}  &44.46 &{36.45} &24.15 &{7.49}  &{6.79}  &{5.83}  \\
 % & AWQ/3.00      &\textbf{6.24}  &5.32  &\textbf{3.74}  &{40.03} &\textbf{43.92} &\textbf{23.59} &8.88  &7.50  &\textbf{6.27}  \\
 % & GPTQ/3.00     &6.44  &5.46  &3.88  &2068.96&61.56 &{18.26} &\textbf{7.95}  &\textbf{7.06}  &5.88  \\
 & GPTQ/2.00     &52.22 &23.63 &8.18  &5583.96&419.07&50.51 &35.27 &19.66 &9.55  \\
 & PB-LLM/1.70   &66.41 &236.40&28.37 &657.24&816.31&N/A   &80.69 &184.67&N/A   \\ 
 & BiLLM/1.08    &32.31 &21.35 &13.32 &5243.01&309.12&72.02 &39.38 &25.87 &17.30 \\
 & ARB-LLM\textsubscript{RC}/1.08    & 16.44 & 11.85 & 6.16 & 389.59& 198.17& 32.79 & 20.38& 14.36 & 8.65 \\
\rowcolor{lightgray!40} & {PTQTP-b1.58-D$^\dagger$}    & \textbf{6.32} &\textbf{5.29} &\textbf{3.95} &\textbf{42.53} &\textbf{46.9} & \textbf{28.15} &\textbf{8.41} &\textbf{7.34} &\textbf{6.3}  \\
\midrule

\multirow{8}{*}{LLaMA-3}
 & FP16    &{6.14}  &N/A   &{2.86}  &{10.05} &N/A   &8.53  &9.54  &N/A   &7.17  \\
 \midrule
 % & AWQ/4.00      &{7.36}  &N/A   &{5.92}  &{11.14} &N/A   &{9.11}  &{10.36} &N/A   &{8.98}  \\
 % & AWQ/3.00      &12.07 &N/A   &10.67 &18.61 &N/A   &27.37 &18.98 &N/A   &22.19 \\
 % & GPTQ/3.00     &18.68 &N/A   &6.65  &18.83 &N/A   &15.97 &17.68 &N/A   &\textbf{10.04} \\
 & GPTQ/2.00     &1480.43&N/A  &82.23 &717.24&N/A   &79.20 &394.74&N/A   &122.55\\
 & PB-LLM/1.70   &73.08 &N/A   &22.96 &106.25&N/A   &45.13 &104.15&N/A   &40.69 \\
 & BiLLM/1.06    &55.80 &N/A   &66.30 &87.25 &N/A   &97.13 &61.04 &N/A   &198.86\\
 & ARB-LLM\textsubscript{RC}/1.06    & 27.42 & N/A & 11.10 & 45.49& N/A & 15.34 & 35.70& N/A& 15.44\\
\rowcolor{lightgray!40} & {PTQTP-b1.58-D$^\dagger$}    & \textbf{8.51} &{N/A} &\textbf{6.12} &\textbf{14.02} &{N/A} &\textbf{10.91} &\textbf{13.24} &{N/A} &\textbf{12.64} \\
\bottomrule
\end{tabular}
% }
\end{table*}

%% file: tables/benchmark_qwen3_14b.tex
\begin{table*}[ht]
% \scriptsize
\small
\setlength{\tabcolsep}{2pt}
\renewcommand{\arraystretch}{1.3}
\centering
\caption{Performance comparison between PTQTP and binary PTQ methods on Qwen3-14B across mathematical reasoning and general benchmarks (accuracy \%). D$^\dagger$: Dual. \textbf{Bold}: best result. }
\label{tab_quanT^14b}
% \resizebox{\linewidth}{!}{%
\begin{tabular}{@{}l*{9}{c}@{}}
\toprule
\shortstack{\textbf{Model}} &
\shortstack{\textbf{Math-500}} &
\shortstack{\textbf{GSM8K}} &
\shortstack{\textbf{ARC-C}} &
\shortstack{\textbf{ARC-E}} &
\shortstack{\textbf{BoolQ}} &
\shortstack{\textbf{HellaSwag}} &
\shortstack{\textbf{PIQA}} &
\shortstack{\textbf{MMLU}} &
\shortstack{\textbf{WinoG}} \\
\midrule
% Min/Max values remain the same.
Baseline-FP16 & 86.60 & 89.39 & 95.25 & 90.83 & 89.45 & 88.09 & 81.07 & 79.38 & 68.90 \\
PBLLM-b1.07
& \applyECTGradedColor{0.00}{0.00}{86.60}{0.00}
& \applyECTGradedColor{1.21}{0.83}{89.39}{1.21}
& \applyECTGradedColor{4.75}{4.75}{95.25}{4.75}
& \applyECTGradedColor{7.76}{4.76}{91.18}{7.76}
& \applyECTGradedColor{29.14}{7.22}{89.45}{29.14}
& \applyECTGradedColor{7.38}{7.38}{88.09}{7.38}
& \applyECTGradedColor{25.46}{7.40}{81.07}{25.46}
& \applyECTGradedColor{12.11}{12.11}{79.38}{12.11}
& \applyECTGradedColor{49.57}{26.44}{68.90}{49.57} \\
BiLLM-b1.05
& \applyECTGradedColor{0.00}{0.00}{86.60}{0.00}
& \applyECTGradedColor{0.83}{0.83}{89.39}{0.83}
& \applyECTGradedColor{50.17}{4.75}{95.25}{50.17}
& \applyECTGradedColor{4.76}{4.76}{91.18}{4.76}
& \applyECTGradedColor{49.39}{7.22}{89.45}{49.39}
& \applyECTGradedColor{22.15}{7.38}{88.09}{22.15}
& \applyECTGradedColor{7.40}{7.40}{81.07}{7.40}
& \applyECTGradedColor{24.40}{12.11}{79.38}{24.40}
& \applyECTGradedColor{49.96}{26.44}{68.90}{49.96} \\
ARB-LLM\textsubscript{RC}-b1.05
& \applyECTGradedColor{1.80}{0.00}{86.60}{1.80}
& \applyECTGradedColor{32.22}{0.83}{89.39}{32.22}
& \applyECTGradedColor{76.95}{4.75}{95.25}{76.95}
& \applyECTGradedColor{82.01}{4.76}{91.18}{82.01}
& \applyECTGradedColor{80.00}{7.22}{89.45}{80.00}
& \applyECTGradedColor{50.01}{7.38}{88.09}{50.01}
& \applyECTGradedColor{69.91}{7.40}{81.07}{69.91}
& \applyECTGradedColor{54.08}{12.11}{79.38}{54.08}
& \applyECTGradedColor{50.36}{26.44}{68.90}{50.36} \\
{PTQTP-b1.58-D$^\dagger$}
& \applyECTGradedColor{82.40}{0.00}{86.60}{\textbf{82.40}} % For bold text, the braces go around \textbf{...}
& \applyECTGradedColor{85.44}{0.83}{89.39}{\textbf{85.44}}
& \applyECTGradedColor{93.56}{4.75}{95.25}{\textbf{93.56}}
& \applyECTGradedColor{88.54}{4.76}{91.18}{\textbf{88.54}}
& \applyECTGradedColor{89.02}{7.22}{89.45}{\textbf{89.02}}
& \applyECTGradedColor{85.12}{7.38}{88.09}{\textbf{85.12}}
& \applyECTGradedColor{79.05}{7.40}{81.07}{\textbf{79.05}}
& \applyECTGradedColor{76.18}{12.11}{79.38}{\textbf{76.18}}
& \applyECTGradedColor{67.88}{26.44}{68.90}{\textbf{67.88}} \\
\bottomrule
\end{tabular}%
% }
\end{table*}